\documentclass[10pt,journal]{IEEEtran}
\IEEEoverridecommandlockouts

\usepackage{mathtools, nccmath}

\usepackage{arydshln}
\usepackage{balance}
\usepackage{multirow,makecell}

\def\BibTeX{{\rm B\kern-.05em{\sc i\kern-.025em b}\kern-.08em
    T\kern-.1667em\lower.7ex\hbox{E}\kern-.125emX}}

\newcommand{\nonl}{\renewcommand{\nl}{\let\nl\oldnl}}
\makegapedcells
\usepackage{cite}
\usepackage{amsmath,amssymb,amsfonts}
\usepackage{makecell}
\usepackage{multirow}
\usepackage[utf8]{inputenc}
\usepackage{enumitem}
\usepackage{graphicx}
\usepackage{textcomp}
\usepackage{subcaption}

\usepackage{algorithm}
\usepackage{algpseudocode}

\usepackage[table]{xcolor}

\usepackage[compact]{titlesec} 
\usepackage{tikz}
\usetikzlibrary{automata,arrows,positioning,calc}

\usepackage{xcolor}
\usepackage{color}
\usepackage{tabularx}
\usepackage{booktabs} 

\DeclareCaptionSubType*[alph]{figure}%
\def\hgr#1{}

\newcommand{\msk}[1]{{\color{purple}#1}}

\newcommand{\RNum}[1]{\uppercase\expandafter{\romannumeral #1\relax}}

\usepackage{pifont}
\usepackage{graphicx}
\usepackage{xcolor}

\newcommand{\RSRP}{\mathit{RSRP}}
\newcommand{\Th}{\mathit{Th}}

\newcommand{\comment}[1]{}
\usepackage[compact]{titlesec}
\titlespacing{\section}{0pt}{0pt}{1pt}
\titlespacing{\subsection}{0pt}{1ex}{0ex}
\titlespacing{\subsubsection}{0pt}{0.5ex}{0ex}

\usepackage{mathtools,etoolbox}
\DeclarePairedDelimiterX{\abs}[1]{\lvert}{\rvert}{\ifblank{#1}{{}\cdot{}}{#1}}

\usepackage{algorithm, algpseudocode}

\begin{document}

\title{
Multi-Agent Deep Reinforcement Learning for Resilience Optimization in 5G RAN}

\author{
{Soumeya~Kaada}, {Dinh-Hieu~Tran},
{Nguyen~Van~Huynh}, {Marie-Line~Alberi~Morel}, {Sofiene Jelassi}, {and Gerardo Rubino}
%
\thanks{The current version of the manuscript is a preprint that was received by the IEEE TMC journal in June 2024 and may differ from the final published version}
\thanks{
S. KAADA is with the Department of computer science, University of Rennes 1- Beaulieu Campus, 35000 Rennes, France (e-mail: soumeya.kaada@etudiant.univ-rennes1.fr).\\
D. H. Tran is with Nokia standards, 91300 Massy, France  (e-mail:  dinh-hieu.tran@nokia.com).\\
N. V. Huynh is with the Department of Electrical Engineering and Electronics, University of Liverpool, Liverpool, L69 3GJ, United Kingdom (e-mail: huynh.nguyen@liverpool.ac.uk).\\
M.L. Alberi Morel is with Nokia Bell Labs, 91300 Massy, France  (e-mail:  marie\_line.alberi-morel@nokia-bell-labs.com).\\
S. Jelassi and G. Rubino are with Irisa and Inria labs, University of Rennes 1- Beaulieu Campus, 35000 Rennes, France (e-mail: sofiene.jelassi@irisa.fr; gerardo.rubino@inria.fr).}
}

\markboth{This manuscript version has been submitted to IEEE Transactions on Mobile  Computing in June 2024 for review.}%
{ }

\IEEEoverridecommandlockouts
\IEEEpubid{\begin{minipage}[t]{\textwidth}\ \\[10pt]
        \centering{Personal use of this material is permitted. Permission from IEEE must be  obtained for all other uses,  in any current or future media, including  reprinting/republishing this material  for advertising or promotional purposes, creating new collective works, for resale or  redistribution to servers or lists, or reuse of any copyrighted component of this work in other works. }
\end{minipage}} 

\maketitle

\begin{abstract}
Resilience is defined as the ability of a network to resist, adapt, and quickly recover from disruptions, and to continue to maintain an acceptable level of services from users' perspective. With the advent of future radio networks, including advanced 5G and upcoming 6G, critical services become integral to future networks, requiring uninterrupted service delivery for end users.
Unfortunately, with the growing network complexity, user mobility and diversity, it becomes challenging to scale current resilience management techniques that rely on local optimizations to large dense network deployments.
This paper aims to address this problem by globally optimizing the resilience of a dense multi-cell network based on multi-agent deep reinforcement learning. Specifically, our proposed solution can dynamically tilt cell antennas and reconfigure transmit power to mitigate outages and increase both coverage and service availability.
A multi-objective optimization problem is formulated to simultaneously satisfy resiliency constraints while maximizing the service quality in the network area in order to minimize the impact of outages on neighbouring cells.
Extensive simulations then demonstrate that with our proposed solution, the average service availability in terms of user throughput can be increased by up to 50-60\%  on  average, while reaching a coverage availability of 99\% in best cases.

\end{abstract}



\section{Introduction}

\IEEEPARstart{T}{he} ever-growing proliferation of mobile users, as well as smart connected devices, has brought a surge in data traffic, coupled with increased complexity in network management. With this exponential growth, the communication network is subject to multiple failures and downtimes.
As 5G and beyond networks become increasingly integral to critical services, the need to optimize their resilience against various threats and uncertainties becomes mandatory for ensuring satisfactory levels of quality of service and user experience. This becomes particularly important with the deployment of many small cells in dense 5G networks, introducing new challenges in resilience and cell outage management~\cite{small_cells},~\cite{small_cell_outage_analysis}. Due to this growing complexity generated by advanced technologies in 5G and beyond networks, and the intensive deployment of network equipment in radio environments that are by nature unstable, the radio access network~(RAN) 
is one of the most critical domains for fault management.

Technically, cell outage management is an essential aspect of resilience management that is widely studied in the self organizing networks (SON), in the context of both cell outage detection and cell outage compensation~\cite{COC_manag}. In fact, an outage can occur due to various reasons such as power failure, disasters, big congestion, and sleeping cell problems. However, detecting the cause of an outage and performing healing mechanisms on it can be a time-consuming task and requires human interventions. For that reason, rapid response through outage compensation for mitigation in various scenarios is needed to ensure service delivery to users from the user's perspective~\cite{COC}. 
In the literature, even though significant work has been done in the area of diagnosis and anomaly detection in troubleshooting~\cite{fault_detection}, outage compensation has not been deeply investigated for the 5G RAN from an experimental approach due to the complex nature of the RAN infrastructure and its associated resource constraints.
Technically, in a RAN, each base station (BS) is assigned to serve a specific area, with little or no redundancy~\cite{RL_COC}.
One potential approach for cell outage management is deploying multiple unmanned aerial vehicles (UAVs) to mitigate outage by covering coverage and capacity gaps during failures~\cite{UAV_COC}. However, UAV's deployment is usually costly and time-consuming.
Another commonly used approach is to adjust antenna neighbouring cell site parameters by activating compensation methods in RAN~\cite{COC},~\cite{COC_effectiveness},~\cite{ COC_adaptive}. For instance, the antenna electrical tilting is a widely adopted technique for enhancing cell coverage and capacity, allowing remote and automatic reconfiguration for network operator engineers~\cite{elec_tilt}. Most of these techniques mainly aim to mitigate the impact of a cell outage by reconfiguring the antenna parameters at each neighbour cell area of the outage BS through local coverage and capacity optimization. The issue of optimizing the coverage and capacity in wireless cellular networks has already been addressed in the literature via off-line optimization approaches~\cite{CCO_literature1},~\cite{CCO_literature2}. Specifically, coverage and capacity are optimized by appropriately tuning the pilot power, antenna tilt and azimuth. While off-line optimization methods may provide valuable insights, it is important to develop methods that adjust involved parameters on-line and in real-time. This is particularly critical with the proliferation of small cell deployments in ultra dense networks~(UDNs), which has increased the complexity of outage management~\cite{small_cells}. In addition, various approaches considered single-objective optimization (e.g., coverage), whereas it has been shown that multiple-objectives optimization (e.g., a combination between coverage and quality of service or interference reduction) must be considered for an efficient performance optimization~\cite{facebook_multi_obj}.

\begin{figure*}[!t]
    \centering
    \includegraphics[width=15cm, height=5.3cm]{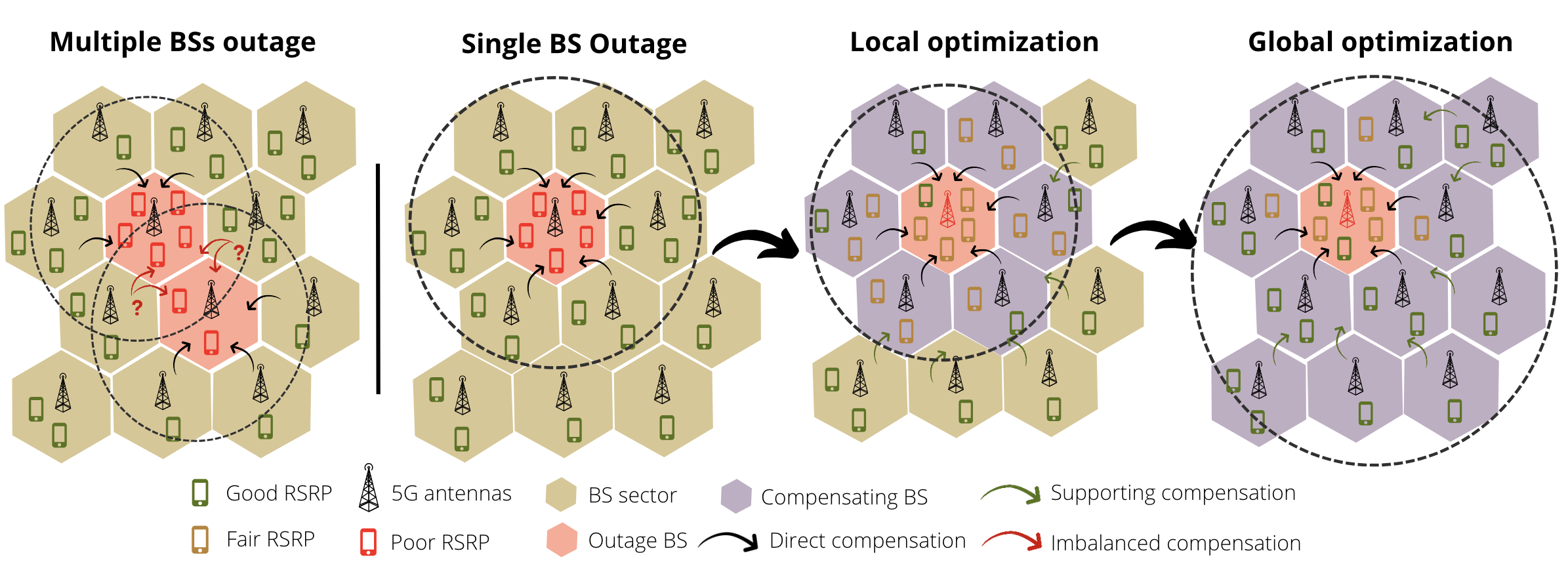}
    \caption{RAN resilience local and global optimization.}
    \label{fig:approach}
\end{figure*}

With the advancement of machine learning (ML), data driven solutions have gained interest in wireless networks for its advantages in learning complex patterns from data. Specifically, deep reinforcement learning (DRL) can autonomously learn optimal decision-making policies, thereby enabling adaptive network management strategies in dynamic and complex environments. Recent studies~\cite{RL_COC},~\cite{RL_COC_dense},~\cite{macro_outage_mitigation}, used DRL for tuning antenna tilt, with either beamforming or transmit power in downlink. However, the majority of the cell outage compensation methods come with a centralized approach involving an outage at one BS while requesting the reconfiguration of only neighbouring sites. Besides, these methods often use local optimizations to scale to large network deployments. 
Moreover, these approaches may not address the potential problems of multiple BS outages, especially when outage BSs are geographically co-located. 
This setting can fail to achieve optimal outage mitigation if more than one outage occurred at the BS, and especially if the failed BSs are located close to each other or adjacent in the considered networking zone. 
In particular, Fig.~\ref{fig:approach} illustrates that the local optimization approach in the case of multiple BSs outage may lead to an imbalanced compensation between the different network outage areas. The neighbouring BSs may not be completely successful in coordinating adjustments between outage areas, leading to a sub-optimal resource allocation and lower network-level performance. Another reason why the conventional approaches based on local optimization may struggle to achieve optimal outage mitigation is the impact on the neighbouring sites and compensating BSs that is not taken into account in previous works. 
This impact was studied in~\cite{impact_BS} by observing degraded metrics on the neighbouring cells. However, the proposition of effective solutions to reduce it during the compensation was missing. In addition, Fig.~\ref{fig:approach} illustrates the case of a single BS outage, where the performance of the neighbouring BSs that attempt to cover the users from the outage zone can be degraded due to an increase in their cell load.

Given the aforementioned issues, we believe that performing a global optimization on the entire network zone can overcome these issues. On the one hand, it will allow us to involve farther BSs to support the direct compensating BS (i.e., neighbouring BSs to the outage area) to absorb the impact and achieve better performance under outage. On the other hand, considering the whole network zone as a cohesive entity 
will automatically balance the optimization during the learning process in cases of multiple BS outages. As such, this article addresses the problem of resilience enhancement based on overall network performance objectives in a multi-cell environment. By leveraging the power of DRL algorithms, we can efficiently address the global optimization and handle the complexity of emerging wireless networks, thereby dynamically adapting and optimizing network resilience in real-time. 
Specifically, we propose to train a multi-agent DRL across an entire network zone. The algorithm will try to optimize the global network performance in terms of coverage and service indicators from the user's perspective under various BS outage situations. To do that, we propose to adjust both the antenna tilt and the transmitted antenna power in the downlink of all cells operating in the considered network service area. 
In~\cite{multi_graph}, the multi-agent DRL was proposed to optimize coverage and capacity in RANs, however, it was not proposed in the context of outage compensation and resilience optimization. In this article,
we develop a novel multi-agent DRL approach to optimize the total user throughput as well as both the coverage and service availability for enhancing the network resilience. For that, we will rely on link budget propagation formulas of the radio environment to compute the estimated reference signal received power~(RSRP) indicator of each user and its correspondent received throughput. This multi-objective optimization allows to achieve an optimal balance for resilience optimization. It priorities ensuring coverage and service availability while improving the quality of service by maximizing the total throughput of the network to reduce outage impacts.
This work is proposed within the scope of extending our previous work on the resilience analysis of a network area~\cite{my_paper1},~\cite{my_paper2}. We claimed that for allowing an end-to-end resilience management, 
an analysis phase of resilience prior to optimization is required to obtain insights about the network performance based on quantifiable metrics. The analysis results will allow to optimize network performance based on those performance metrics. This enables operators to implement targeted optimization strategies based on their analysis and treat the network as a cohesive entity to launch an automated resilience management scheme 
on demand.

 
In this article, we deploy the distributed setting of our proposed multi-agent DRL within a small cell UDNs assuming that outage occurs 
when the network performance is below the requirements using the resilience analysis approach developed in our previous work~\cite{my_paper2}. We are interested in a scenario where one or multiple BSs can experience an outage. We perform experiments with various settings and scenarios using a standardized simulator developed by Nokia~\cite{SONTool1}. Unlike previous works such as~\cite{RL_COC},~\cite{RL_COC_dense},~\cite{macro_outage_mitigation}, we will rely on realistic radio channel characteristics and standardized propagation formulas. Extensive simulations reveal that our proposed solution can increase service availability by up to 50-60\% on average, while reaching a coverage availability of 99\% in best cases.
In summary, our contributions are as follows:

\begin{itemize}
    \item We consider a multi-cell network and aim to dynamically and intelligently optimize its resilience. In particular, we first formulate a multi-objective optimization problem to satisfy the coverage and service availability requirements and at the same time improve quality of service by maximizing the total throughput to reduce the impact of outage compensation on the compensating BSs. This is achieved by requesting the reconfiguration of not only neighbouring BSs of the outage area but also farther BSs within the considered network zone to play a role in supporting the compensation (see Fig.~\ref{fig:approach}).

    
    \item Given that the formulated optimization problem is non-convex and NP-hard, even when all parameters of the system are known in advance, implementing them in practice is not feasible. Hence, we propose a novel multi-agent DRL for optimizing the resilience of a 5G RAN area using global optimization to scale to large network deployments. The distributed setting of the proposed DRL method considers the network as a cohesive entity to capture 
    its complexity and the relations between cells.
    Specifically, we consider the joint configuration of both antenna transmit power and antenna downtilt for all the operational BSs in the considered area to ensure an optimal coverage and capacity optimization (see Fig.~\ref{fig:approach}).

    \item We take into account different outage scenarios in the training process to allow our proposed multi-agent DRL algorithm to face various outage situations. This addresses one of the key challenges of resilience optimization which is enhancing resilience regardless of the unpredictable nature of threats and uncertainties that networks will encounter.

    \item We validate the effectiveness of our proposed approach on a standardized simulator, on which the numerical results show significant improvements in both coverage and service resilience as well as optimal quality of service enhancement compared to conventional approaches. For that, realistic propagation formulas and standardized path loss of small cell deployment in UDNs were used to practically demonstrate the usability of our proposal. 

    \item This work is developed in the context of enabling an end-to-end resilience management strategy.
    More specifically, our proposed approach extends our previous works on resilience analysis. It integrates resiliency constraints extracted from the analysis results in the optimization process. By linking resiliency analysis and its optimization within a network area, it enables the activation of an automated resilience management strategy. This strategy dynamically adapts the network to efficiently deal with various outage scenarios, following the targeted needs of network operators.

\end{itemize}




The rest of the paper is structured as follows. {Section~\RNum{2}} describes in detail the radio system and propagation environment as well as the radio data used during the resilience management process.
{Section~\RNum{3}} briefly reviews the resilience analysis approach and its role in the optimization. In {Section~\RNum{4}}, we develop our resilience optimization problem and the proposed multi-agent DRL algorithm. In {Section~\RNum{5}}, we present our network setups, the training performance, and the validation of our proposed approach in the simulator. Finally, conclusions are given in {Section~\RNum{6}}.





\section{System description}
In this section, we start by describing the radio user key performance indicators (KPIs) considered during the resilience optimization and evaluation. Then, we detail the propagation environment and the link budget formulas.

\subsection{Data description}

A UE is  able  to  connect  to  its  mobile  network if it receives a signal strength from BS above an acceptable threshold. This threshold is defined based on the antenna receiver sensitivity. Additionally, the user should be able to establish and maintain a  session  for  a  minimum  period  of  time. To maintain a good connection for a user and access a service, the UE should achieve a specific data transmission speed (i.e., throughput  or  data  rate)  over  the  established connection. Other quality of service indicators are also important such as latency to be satisfied but they are out of the scope of this work.

In 5G, the channel state information~(CSI) - radio received signal strength $\RSRP$ is a valuable layer-1 measurement to provide information about the signal strength received by users. It is defined as a linear average over the power contributions (in Watts) of a single reference signal (RS) resource element (in dBm). The $\RSRP$ can  act  as  an  indicator  of  the radio coverage within a service area. The classification of $\RSRP$ levels shown in Table~\ref{table:RF_conditions} facilitates the categorization of network conditions and their associated degradation.
Besides, Throughput ($\Th$) in Kbps, gives indication about the quality of service of the user session. It depends on the bandwidth, channel quality and the resource blocks~(RBs) allocated. The $\RSRP$ indicator reflects the signal strength received by users and the $\Th$ reflects its correspondent data rate. During network planning, these indicators are estimated based on the link budget propagation formulas detailed in the next section.

\begin{table}[!h]
\small
\centering
\begin{tabular}{ |c|c|c|} 
 \cline{2-3}
 \multicolumn{1}{c|}{} & Classification   & \cellcolor{orange!80} \textbf{ $\RSRP$ (dBm)}  \\ [0.5ex] 
 \hline
  \cellcolor{orange!80} &  &  \\ 
  
   \cellcolor{orange!80} & Good (or excellent) 
   & -90 to -44   \\ 
 
  \cellcolor{orange!80}& Fair (Mid cell) 
  & -126 to -91 \\ 

  \parbox[t]{10mm}{ \cellcolor{orange!80}\multirow{-5}{5pt}{ \rotatebox[origin=c]{90}{ \thead{\small RF \\ \small  Conditions}  }}} & Poor(Cell Edge) 
  & -140 to -127 \\ [6pt] 
 \hline
 \end{tabular}
 \caption{Radio RF condition configuration ($\RSRP$).}
 \label{table:RF_conditions}
\end{table}


\subsection{Propagation Environment}

We  consider  a  mobile  network  service  area  in  an  urban environment served by multiple BSs.
Let us  assume  that  each BS site is equipped with three antennas, each covering one sector. From a practical perspective, this assumption corresponds to typical mobile network deployments as a BS serves users inside its sector. It is worth noting that, the term cell is used to refer to one sector in this paper. We assume that each sector can be geometrically approximated as a hexagon, thus the  network is multi-cellular  and  sectorized. In  practice,  due to  inconsistent  radio  propagation characteristics and environment landscape, the shape of the sector is not a perfect hexagon~\cite{jointoptjournal}.

Let $M$ denotes the total number of cells in a given geographical area and $N$ denotes the number of UEs in the same network area at specific time $t$. A UE connects to the network through its Mobile Station (MS) or User Terminal (UT) and is served by a specific BS in the given service area. 
We suppose, that the $j^{th}$ BS antenna, with~$j \in \{1, \ldots, M\}$ and~$M >> 3$, transmits signal with power $P_{t_j}$ in the network area. The signal is amplified with the gain of the BS antenna, denoted as $G_tj(\theta_j)$ where $\theta_j$ is the antenna downtilt. The signal propagates through the wireless channel between the $j^{th}$ BS antenna and the $i^{th}$ UE, where $i \in \{1, \ldots, N\}$.
Radio signals launched by a transmitter experience signal attenuation as they traverse through the propagation channel. This attenuation is a function of the carrier frequency, the heights of the BS and MS, the BS-MS link distance $d$, as well as the environment-type (i.e., dense urban, suburban or rural). Path loss is a large-scale fading parameter, which determines the mean signal attenuation as a function of the BS to MS distance.

\textit{\textbf{Path loss:}}
In contrast to conventional approaches that often rely on the simplified free-space path loss model described by Friis' propagation formula for its simplicity~\cite{free_space_pathloss},
this study adopts a more realistic approach by considering a real-world network deployment scenario. Specifically, we focus on an urban environment and leverage the standardized 3GPP path loss model tailored for small cell deployments known as the urban micro~(UMi) path loss model, as defined in the 3GPP specification~TR~38.901~\cite{5G_channel_standard}. By utilizing this path loss, we aim to better capture the complexity of signal propagation in urban environments. This approach allows us to account for factors such as building density, street layouts, and other urban features that significantly influence signal propagation characteristics. It will provide a more accurate representation of the propagation environment, leading to more reliable and informative results for network planning and optimization. 

Assuming that both the BS and MS are separated by a distance $d_{2D}$ and $d_{3D}$ in meters (m) in both 2D and 3D planes, respectively. 
A BS with height $h_{BS}$ transmits a signal with frequency $f$ (corresponding to wavelength $\lambda$), to the UT with a height of~1.5m$\leq h_{UT} \leq$22.5m. The correspondent Line Of Sight~(LOS) path loss equations in dB scale described in~\cite{5G_channel_standard} as follows:

\begin{equation}
\begin{aligned}
  PL_{UMi-LOS} = 
  \begin{cases}
  PL1 & \text{if } 10m\leq d_{2D} \leq d'_{Bp}, \\
  PL2 & \text{if } d'_{Bp}\leq d_{2D} \leq 5km,
  \end{cases}
\end{aligned}
\label{eq:UMi_pathloss}
\end{equation}
where,
\[
\begin{aligned}
PL1 &= 32.4 + 21 \log_{10}(d_{3D}) + 20 \log_{10}(f_c), \\
PL2 &= 32.4 + 40 \log_{10}(d_{3D}) + 20 \log_{10}(f_c) \\
& - 9.5 \log_{10}((d'_{Bp})^2 + (h_{BS}-h_{UT})^2).
\end{aligned}
\]
Here, the number $d'_{Bp}$ is the Breakpoint (Bp) distance $d'_{Bp}~=~4 h'_{BS} h'_{UT} f_c/c$, 
where the effective BS antenna and UT antenna heights are computed respectively as follows: $h'_{BS} =  h_{BS} - h_e$ and $h'_{UT} =  h_{UT} - h_e$. In the case of UMi environment, the effective environment height is $h_e = 1m$. Besides, $f_c$ is the carrier frequency of the transmitted signal in Hz, and the light speed is~$c = 3 \times 10^8~m/s$. For the sake of simplicity the non line of sight path loss equation $PL_{UMi-NLOS}$ is not showed here, but it is described in 3GPP standard~\cite{5G_channel_standard}, as well as in the standardized simulator used in his work.

\begin{figure}[!hbp]
    \centering
    \includegraphics[width=8.5cm, height=3.5cm]{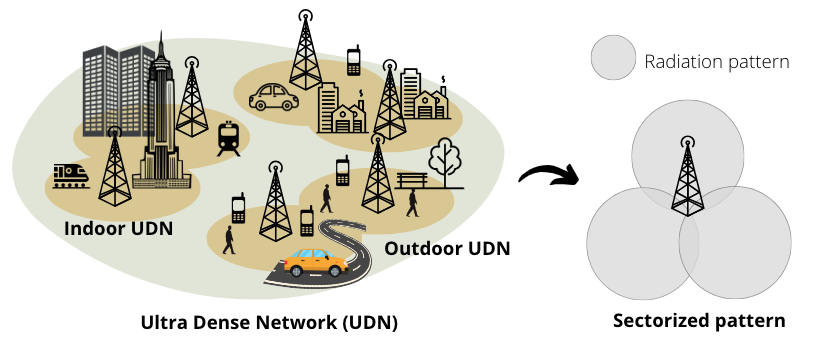}
    \caption{UDNs and antenna radiation pattern.}
    \label{fig:beamforming_antenna}
\end{figure}

\textit{\textbf{Received signal:}}
When the $j^{th}$ BS antenna transmits a power $P_{t_j}$, it is first amplified by the gain of the BS antenna before it faces attenuation in the air interface characterized by the path loss equations $PL_{UMi}$, where (\ref{eq:UMi_pathloss}) describes the LOS path loss formula. The received user signal strength $P_r$ is defined as: 
 

\begin{equation}
    P_r = P_{t_j} + G_tj(\theta_j) + G_r - PL_{UMi},
    \label{eq:received_signal}
\end{equation}
where $G_tj(\theta_j)$ is the $j^{th}$ transmit BS antenna gain expressed in dBi and $\theta_j$ is the antenna downtilt. The formula of the BS antenna gain is further described in~(\ref{eq:gain_t}). Besides, $G_r$ denotes the received user antenna gain in dBi. It is usually set as~0 dB for an omnidirectionnal antenna like broadband patch antennas used in mobile phones since they radiate equally in all directions like an isotropic antenna (i.e.,~0 dBi equivalent to~0~dB since the reference gain in 5G that is added to the antenna gain is~0~dBi to obtain the gain in dB scale). Finally $PL_{UMi}$ is the path loss described in~(\ref{eq:UMi_pathloss}). A user is capable to connect to a BS cell if he receives a signal strength above the minimal threshold $Pr_\epsilon$. This threshold is related to the antenna receiver sensitivity, and is defined to be around~-127dBm as shown in Table~\ref{table:RF_conditions}.

\begin{figure}[!htp]
    \centering
    \includegraphics[width=7cm, height=3.5cm]{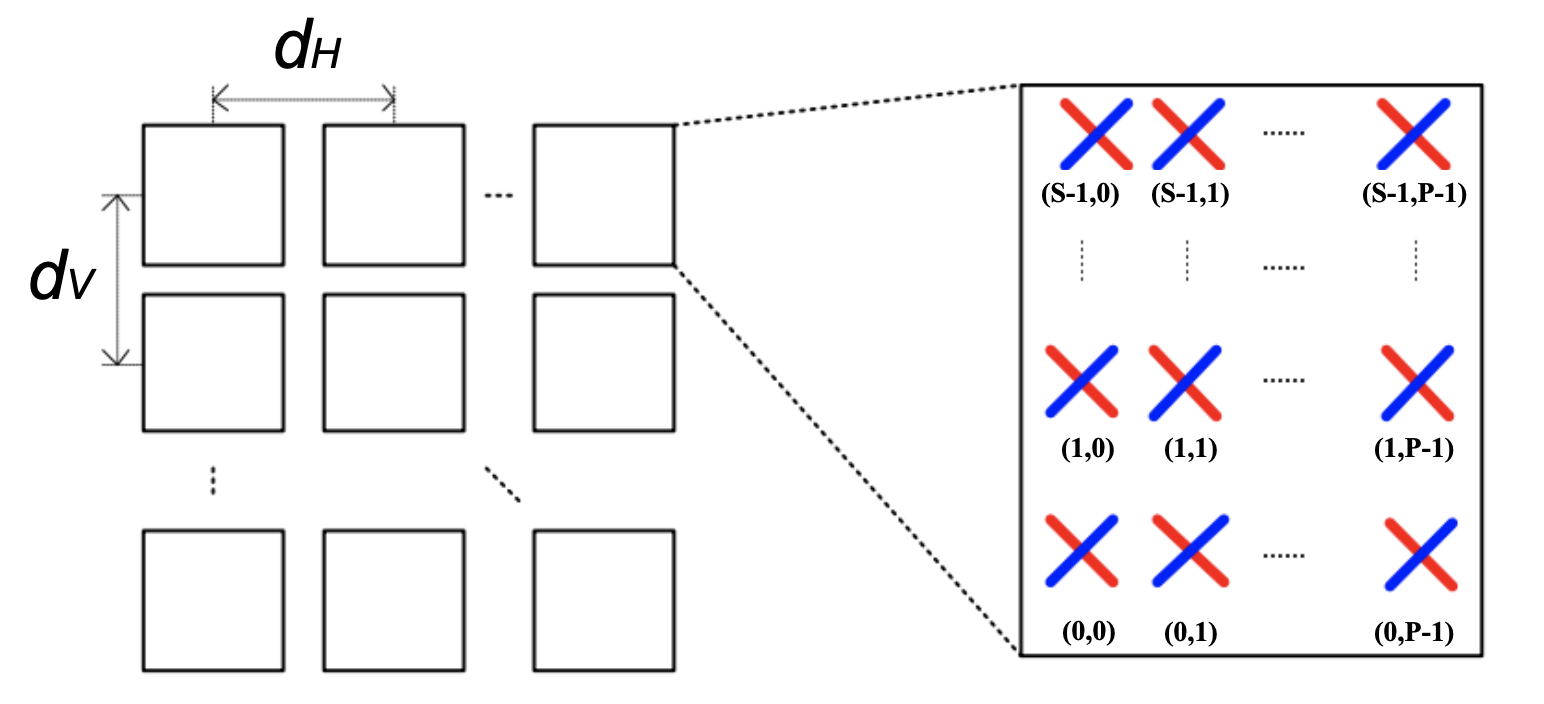}
    \caption{Cross-polarized antenna panel~\cite{5G_3DBF_mmwave}.}
    \label{fig:cross_polar}
    \vspace{-0.4cm}
\end{figure}

\begin{figure*}[!htp]
\centering
    \begin{subfigure}[t]{0.45\textwidth}
    \centering
    \includegraphics[width=11cm, height=6cm]{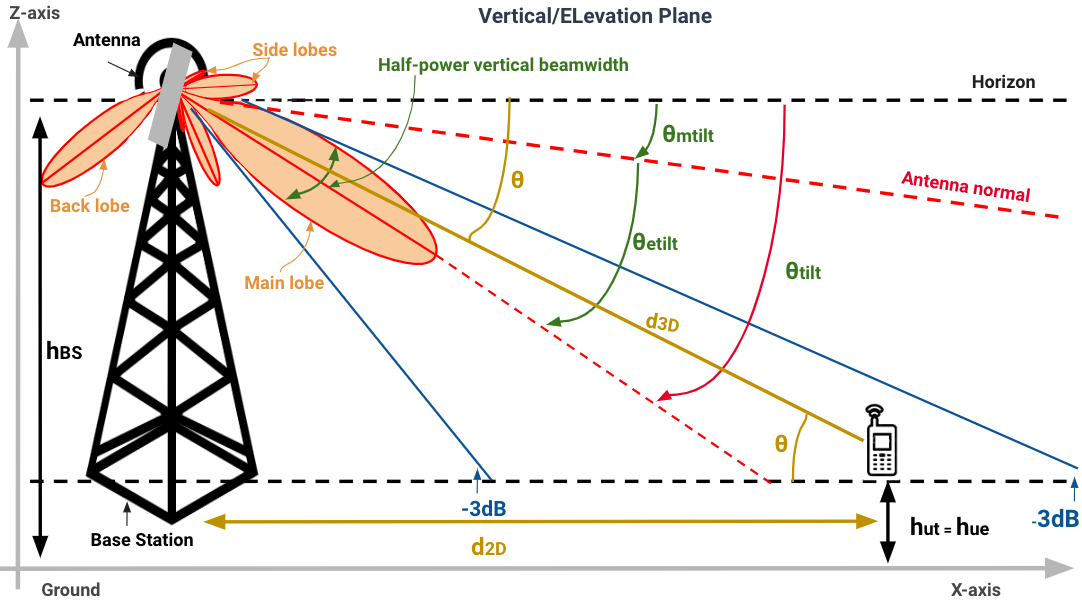}
    \caption{}
    \end{subfigure}
    \hspace{0.5em}
    \begin{subfigure}[t]{0.45\textwidth}
    \raggedleft
    \includegraphics[width=5.5cm, height=5cm]{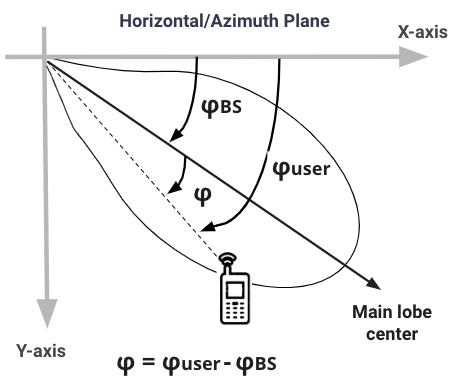}
    \caption{}
    \end{subfigure}
    \caption{Transmission pattern parameters: (a) Elevation plane and (b) Azimuth plane.}
    \label{fig:transmission_pattern}
\end{figure*}

\textit{\textbf{Antenna radiation Pattern:}}
In 5G BSs deployment, a 3D antenna pattern is usually used in combination with cell sectorization to exploit frequency reuse, reduce interference among users, and increase cell capacity (see Fig.~\ref{fig:beamforming_antenna}). In this method, each sector is served by an array of~$(S \times P)$ cross polarized antennas as shown in Fig.~\ref{fig:cross_polar}. The BS antenna is modelled by a uniform rectangular panel arrays
spaced in the horizontal direction with a spacing of~$d_h$ and in the vertical direction with a spacing of~$d_v$~\cite{5G_channel_standard}. For the sake of simplicity, we assume having a one dimensional antenna vertical array panel~($P=1$) at each cell sector.
The 3D antenna pattern is typically given by a horizontal
(azimuth) and a vertical (elevation) cuts as shown in Fig.~\ref{fig:transmission_pattern}. This extrapolation is typically used by 3GPP to approximate the real 3D pattern.
We assume that user $u$ having a height $h_{UE}$ from the ground, is served by cell $j$ with a height $h_{BS}$. Every user is connected to exactly a single BS and is served by its correspondent antenna serving cell $c$. The angles formed by the transmitted beam from the BS are extrapolated in Fig.~\ref{fig:transmission_pattern} in both vertical (elevation) and horizontal (azimuth) planes. Having an omnidirecrtionnal UE’s  antenna pattern, the angles of arrivals $\theta$ and $\phi$ between the BS antenna main lobe  and the UT in elevation and azimuth planes respectively, are defined by the following equations:

\begin{equation}
    \theta = \tanh^{-1}{\frac{h_{BS} - h_{UE}}{d_{2D}}},
    \label{eq:angle_user_theta}
\end{equation}

\begin{equation}
    \phi = \tanh^{-1}{(\phi_{user} - \phi_{BS})},
    \label{eq:angle_user_phi}
\end{equation}
where $d_{2D}$ is the distance between the BS antenna and the UT in 2D and $\Phi_{BS}$ is the azimuth orientation of the BS antenna. Typically a single BS serves three cells (i.e., sectors) which have the azimuth orientation $\Phi_{BS}$ = \{0°, 120°, -120°\}~\cite{SONTool1}.

We assume having a sectorized antenna radiation pattern in the azimuth $A_v$ and elevation $A_h$ planes as defined in~\cite{LTE_sector_gain}. We follow the extrapolation approach in~\cite{jointoptjournal},~\cite{SONTool1}, that sum up the two patterns to get the 3D pattern $G(\theta, \phi)$ as follows:

\begin{equation}
    A_v(\theta) = - \min [12 \;\;(\frac{\theta -\theta_{tilt}}{\theta_{3dB}})^2, SLL_v],
    \label{eq:vertical_pattern}
\end{equation}

\begin{equation}
    A_h(\phi) = - \min [12 \;\;(\frac{\phi}{\phi_{3dB}})^2, SLL_h],
    \label{eq:horizontal_pattern}
\end{equation}

\begin{equation}
    G(\theta, \phi) = G_{max} - [A_v(\theta) + A_h(\phi)],
    \label{eq:gain_t}
\end{equation}
where $\theta$ and $\phi$ are the angles of arrival between the BS antenna main lobe  and the UT receiver defined in equations (\ref{eq:angle_user_theta}) and (\ref{eq:angle_user_phi}) in the Cartesian referential. $\theta_{tilt}$ designates the antenna downtilt angle with $\theta_{tilt} = \theta_{mtilt} + \theta_{etilt}$. The mechanical tilt is usually fixed and can be modified manually by human intervention, whereas the electrical tilt can be modified remotely by tuning the transmit signals phases~\cite{net_elec_tilt}. 
When~$\theta_{tilt} =0$, the antenna points to the horizon.
$G_{max}$ is the maximum antenna element gain that is usually specified between~7-15dBi by the manufacturer. Moreover,~$\theta_{3db}$ and $\phi_{3db}$ represent the~3dB beam-width in the elevation and azimuth planes.
Finally, $SLL_v$ and $SLL_h$ are the azimuth and elevation side lobe levels respectively, that provide a minimum power which is leaked to the sectors other than the desired one and which have a typical value of 15 to 30 dB.



\textit{\textbf{Throughput Calculation:}} The data rate (i.e., user Th) with which the UE is served by a cell antenna can be expressed using the Shannon–Hartley theorem formula based on the number of physical resource block (PRB) allocated to the user
in bits per second as follows: 


\begin{equation}
    Th = B_{PRB} \cdot N_b \cdot \eta \cdot \log_{2}(1+\frac{Pr}{I + n_0}),
    \label{eq:throughput}
\end{equation}
where $B_{PRB}$ represents the bandwidth of each PRB and $N_b$ is the number of bits per symbol in each PRB. Besides, $\eta$ is the number of PRBs allocated to the user which is computed using the power unit of the UT device and the available number of PRBs at the BS antenna. The throughput is also computed using the Shannon spectral efficiency $\log_{2}(1+\frac{Pr}{I + n_0})$, where the fraction $\frac{Pr}{I + n_0}$ represents the signal-to-noise-and-interference ratio. The ratio is expressed as the factor between the received signal $Pr$ and the thermal noise $n_O$  with the interference power $I$. The thermal noise is usually considered as the additive white Gaussian noise (AWGN) with~$n_0 \sim \mathit{N}(0, \sigma^2)$ representing a Gaussian random variable with zero mean and variance $\sigma^2$. Besides, $I$ represents the total power of interfering signals received at the user and transmitted by the non serving cells, except the power signal received by the serving cell $c$, such that $I = \sum_{\substack{j\in M \\ j \not = c}}^{}Pr_j $ in watt scale. A condition is established on the throughput, affirming that if a user can not be covered by a BS then the user is unable to access data services. This means that when $Pr < Pr_\epsilon$ then $Th = 0$, emphasizing the role of coverage in enabling service accessibility. We summarize the parameters used in this paper in Table~\ref{table:list_parameters}.

\begin{table}[htp]
  \centering
  \begin{tabular}{!{\vrule width 1pt}p{3cm}|p{5cm}!{\vrule width 1pt}} 
    \specialrule{0.1em}{0em}{0em}
    \textbf{Parameter} & \textbf{Value} \\
    \specialrule{0.1em}{0em}{0em} 
    $\theta$, $\phi$ & Angles of Arrivals (AoA) between BS antenna main lobe beam and the UE terminal \\
    $\theta_{tilt}$ & antenna downtilt in elevation pattern \\
    $\theta_{mtilt}$,  $\theta_{etilt}$ & antenna mechanical and electrical downtilt in elevation pattern \\
    $SSL_v$, $SSL_h$ & Side Lobe levels of vertical and horizontal patterns respectively\\
    $\phi_{BS}$ & Boresight angle of the antenna BS \\
    $A_v$, $A_h$ & antenna radiation patterns in vertical (elevation) and horizontal (azimuth) planes respectively \\
    $\theta_{3dB}$, $\phi_{3dB}$ & Half power beamwidth of the elevation and azimuth patterns, respectively\\
    $G_{max}$ & Maximum gain of the BS antenna\\
    $G(\theta,\phi)$ & 3D BS antenna gain \\
    $h_{BS}$, $h_{UT}$ & Heights of Base Station and UT from the ground\\
    $G_r$ & Received antenna gain of UE \\
    $PL_{UMi}$ & Path loss function of the urban micro cells area\\
    $P_t$ & Transmit antenna signal strength in dB scale\\
    $d_{2D}$, $d_{3D}$ & distance between antenna BS and UE terminal in 2D and 3D respectively\\
    $d_ij$ & Distance between $j^{th}$ antenna BS and $i^{th}$ UE terminal\\
    $L$ & Number of outage Base Stations \\
    $UE_i$, $BS_J$ & $i^{th}$ UE terminal and $j^{th}$ antenna BS\\
    $Pr_\epsilon$ & Minimal threshold of the received signal required to cover the user\\
    $D_\epsilon$ & Minimal requested throughput by the $i^{th}$ user\\
    \specialrule{0.1em}{0em}{0em}
  \end{tabular}
  \caption{List of all parameters.}
  \label{table:list_parameters}
  \vspace{-0.6cm}
\end{table}

\section{Resilience analysis and its optimization}

This study is an extension to our previous work on resilience analysis based on continuous time Markov chains (CTMCs) modelling. We refer interested readers to~\cite{my_paper2} to find more insights about quantifying the resilience of a RAN area based on coverage and quality of service indicators. Nevertheless, in this section we will review briefly
our previous work on resilience analysis
for coverage and service that use $\RSRP$ and user $\Th$ indicators data respectively. This analysis will be considered in the resilience optimization algorithm and used for its evaluation in the experimental section.


\subsection{Resilience Analysis Review}
\label{sec:analysis}

To analyse the network resilience, we use two CTMCs to model the different network states of the $M$ cells in a given geographical area. 
Let $X_{\RSRP}(t)$ (see Table \ref{table:RF_conditions}) and $X_{\Th}(t)$ be the CTMC models that capture radio coverage and service integrity states using $\RSRP$ and $\Th$ of all users, respectively. Both models have 5 states with full transitions. 
Due to their similarities, we define the generic CTMC model~$X(t)$  with finite state space $S = \{G = ``Good'', F = ``Fine'', A = ``Acceptable'', P = ``Poor'',  O = ``Outage''\}$. 

\textbf{(a) Radio coverage:} We define each state of a coverage area in terms of the tuple $(x,y,z)$, as respectively the percentage of users receiving good, fair, and poor $\RSRP$ levels as shown in Table~\ref{table:RF_conditions}. We set out to~$5\%$ the admissible percentage of users, that receive an $\RSRP$ below the threshold~$Pr_\epsilon= -127dBm$ associated with poor $\RSRP$. 
This implies that the network operator should guarantee a minimal radio coverage for 95\% of users (i.e., $\hat{P}_{coverage} = 0.95$) to be resilient from coverage perspective. It is directly inspired from assumptions of mobile network radio planning. Thus the states definition for coverage are:

\begin{enumerate}

    \item $X_{RSRP}(t) = G$ for \textit{``Good''} if $x > y$, $x \geq z$, $x > y + z$ and if the percentage of users with poor $\RSRP$ verifies $z < 5\%$. 
    
    \item $X_{RSRP}(t) = F$ for \textit{``Fine''} if $x > y$, $x \geq z$ and the percentage of users with poor $\RSRP$ verifies $z < 5\%$ but $x \leq y + z$. 
    
    \item $X_{RSRP}(t) = A $ for \textit{``Acceptable''} if $y > x$, $y \geq z$ and the percentage of users with poor $\RSRP$ verifies $ 0 \leq z < 4\% $.
    
    \item $X_{RSRP}(t) = P $ for \textit{``Poor''} if $y>x$, $y\geq z$ and the users percentage with poor $\RSRP$ verifies $ 4\leq z<5\%$.   
    
    \item $X_{RSRP}(t) = O$ for \textit{``Outage''} if the percentage of users having poor $\RSRP$ verifies $z \geq 5\%$.
    
\end{enumerate}

This means that cells carry traffic with best (respectively fine, acceptable) performances in state $G$ (respectively $F, A$). In $P$, they carry some traffic, but with poor performance, and in $O$, the coverage is at least severely degraded, and possibly with no traffic carried. The reader can verify that all states are fulfilled by the tuple $(x,y,z)$ where all percentage sum is 100\%, and the coverage status of the service area can only have one state at a time. Based on the states definition and the minimal coverage condition,  
we consider that the radio system is resilient from coverage perspective at time~$t$ if~$X_{RSRP} =~G$,~$F$,~$A$,~or~$P$. The coverage level threshold $\hat{P}_{coverage}$ can vary depending on the operators needs, and the definition of network coverage states can also be customized.

\textbf{(b) Service integrity:} We define each state in terms of users' percentage $d$ that receives at least the requested data rate threshold $D_\epsilon$ by each user, namely $\Th \geq D_\epsilon$. 
The states are defined as follows:

\begin{enumerate}

    \item  $X_{Th}(t) = G$ for \textit{``Good''} if $d \geq 80\%$.
    \item  $X_{Th}(t) = F$ for \textit{``Fair''} if $65\% \leq d < 80\%$.
    \item  $X_{Th}(t) = A $ for \textit{``Acceptable''}  if $50\% \leq d < 65\%$.
    \item  $X_{Th}(t) = P $ for \textit{``Poor''} if $30\% \leq d < 50\%$.
    \item  $X_{Th}(t) = O$ for \textit{``Outage''}  if $d < 30\%$.
    
\end{enumerate}

We define network resilience from service perspective as the ability of the network to fulfill the minimal throughput requirements for 50\% of users (i.e., $\hat{P}_{service} = 0.5$).
Thus, we consider that the radio system is resilient from service perspective at time $t$ if $X_{Th} = G$, $F$, or $A$. The service level threshold $\hat{P}_{service}$ can vary depending on the operators needs, and the definition of service states can be customized.

\subsection{Resilience analysis role in the optimization:}
Using Markov chains to model the network into multiple states allows us to represent the network status in terms of levels for both coverage and quality of service. Based on the Markov models, various metrics can be derived to quantify network resilience. In the literature, researchers have defined resilience metrics as the availability of network services over a given time period~\cite{my_paper2},~\cite{ilaria_paper}. These metrics aim to characterize the network's 
resilience, which is defined as the ability of a network to maintain its services and operational functionalities under failures. However, most online optimization approaches focus on enhancing resilience at specific time instances, thereby indirectly improving resilience over the long term (i.e., over a period of time). Consequently, the conditions used to define each resilience level in $X_{RSRP}$ and $X_{Th}$ (i.e., coverage and service level thresholds $\hat{P}_{coverage}$ and $\hat{P}_{service}$) can serve as constraints in the optimization objective function.
We claim that the iterative process of resilience optimization at time~$t$, contributes positively to long-term network performance, 
thereby improving metrics of resilience over time.

Considering these conditions as constraints, ensures that the optimization process takes into account the desired coverage and quality of service levels under different circumstances, following a tailored and multi-objective optimization.
Moreover, the resilience analysis offers a clear way for evaluating the resilience optimization and how the network performance of both coverage and quality of service evolves under various conditions and scenarios. Therefore, aligning the resilience analysis with its optimization will provide an end-to-end resilience strategy.
To explore the implementation of this strategy within a RAN, we refer the reader to our previous work on enabling a resilience management framework at RAN level~\cite{my_paper1}.



\section{Multi-agent DRL for resilience optimization}

In this section, we detail our solution by giving the problem formulation of the multi-objective optimization of resilience and model it using Markov decision process~(MDP). Then, we present the multi-agent DRL algorithm for resilience optimization.

\subsection{Problem formulation}
\label{sec:probb_form}

We consider a service area with~$M$~BS and~$N$~test user points defined by their coordinates, where the user points coordinates are $UE_i~\in~\{(x_1,y_1),\cdots, (x_N,y_N)\}$, and the BSs coordinates are $BS_j~\in~\{(a_1,b_1),\cdots, (a_M,b_M)\}$. The distance between the $i^{th}$ user and the $j^{th}$ BS antenna in 3D is given by $d_ij = \sqrt{(x_i-a_j)^2 + (y_i-b_j)^2 + (h_{BS}-h_{UE})^2}$.

Let us assume that a subset of~$L$~BSs are in an outage state due to various reasons not considered in this study, such that $BS_O =\{BS_{o_1},..,BS_{o_L}\}$ where $L < M$.
We are interested in a scenario where a certain number of BSs are off, making some users to loss the connection and degrade their service. Thus, the status of the network reflects a problem within users KPIs.
We tune both antenna electrical downtilt $\theta_{etilt}$ and transmit power $P_t$, to optimize the users KPIs that are the user throughput $Th$ and the received signal $Pr$ in addition to coverage and service availability defined in section~\ref{sec:analysis} at time $t$, such that, the resilience of the network measured over a period of time is also improved.

Let $\theta_{etilt_j} \in [0^\circ,14^\circ]$ be the set of possible tilt values of functional BSs and $P_{t_j} \in [0dB,40dB]$ the set of possible values of the transmit antenna power of these same BSs. Unlike related studies that tune only neighbouring antennas sites radio parameters, we attempt to tune both tilt and transmit power of all functional BSs belonging to the set $\{BS\}\smallsetminus \{BS_O\}$ within the network area (see Fig.\ref{fig:approach}). The goal is to ensure, in the one hand, that coverage and service are consistently available across the service area, and, in the other hand, that the network is reducing the compensation impact in the area by delivering higher data rates and better quality of service, resulting in solving a multi-objective function. To define coverage and service availability metrics in the optimization problem, we first define the following binary variables:

\begin{equation}
    \alpha_i = 
    \begin{cases}
      1, & \text{if}\ Pr_i \ge Pr_\epsilon, \\
      0, & \text{otherwise},
    \end{cases}
    \label{eq:coverage_availability}
\end{equation}

\begin{equation}
    \beta_i = 
    \begin{cases}
      1, & \text{if}\ Th_i \ge D_\epsilon, \\
      0, & \text{otherwise},
    \end{cases}
    \label{eq:service_availability}
\end{equation}
where $Pr_\epsilon$ is the threshold of the received signal in dBm (i.e., $\RSRP$) required to connect a user to a serving BS and $D_\epsilon$ is the minimal requested data rate by the user, both defined in section~\ref{sec:analysis}. Thus the availability of resilient coverage and service are defined as: 

\begin{equation}
    P_{coverage} =  \frac{\sum_{i}^{N}{\alpha_i}} {N}, \quad P_{service} =  \frac{\sum_{i}^{N}{\beta_i}} {N}
    \label{eq:P_c_and_P_s},
\end{equation}
Following the analysis of resilience and constraints levels defined in section~\ref{sec:analysis}, we consider that the system is resilient at time $t$ if $P_{coverage} \ge \hat{P}_{coverage}$ and $P_{service} \ge \hat{P}_{service}$. For that, we define a binary variable that verifies either the network is satisfying the coverage and service constraints or not as follows:

\begin{equation}
    z = \begin{cases}
      1, &  \begin{array}{l} 
            \text{if } \mathit{P_{coverage}} \ge \mathit{\hat{P}_{coverage}} \\ 
            \text{and } \mathit{P_{service}} \ge \mathit{\hat{P}_{service}} 
          \end{array} 
      \\ 
      0, & \text{otherwise}.
    \end{cases}
    \label{eq:qos_availability}
\end{equation}

Under the hypothesis of fixed traffic, we formalise the optimization problem in~(\ref{eq:general_problem}) that represents a linear combination
between maximizing network throughput $\Th$ (see~(\ref{eq:throughput})), and ensuring network coverage and service availability. The optimization problem~(\ref{eq:general_problem}) ensures the continual improvement in quality of service when the resilience thresholds are met, leading to a reduction in the impact of outage on compensating BSs. In the case where the resilience constraints fall under the desired thresholds, the second part of the optimization function is formulated to persist in maximizing coverage and service availability. This is intended to continue mitigating outage and accommodate the maximum number of users across the service area. Consequently, this optimization problem is designed to balance between maximizing throughput and ensuring coverage and service availability, thereby enhancing overall network performance and resilience. We formalise our problem as follows:

\begin{equation}
\begin{aligned}
\max \quad & z \sum_{i=1}^{N}{Th_i} + (1-z) \; P_{coverage} \times P_{service} \\
\textrm{subject to} \quad & C1: \; \rm{if} \; \mathit{P_{coverage}}\ge \mathit{\hat{P}_{coverage}} \;\;  \\ &  \;\;\;\;\;\;\;\;\rm{and} \;\;  \mathit{P_{service}} \ge \mathit{\hat{P}_{service}} \implies z = 1,  \\
& C2: \; Th_i \ge D_\epsilon \implies \beta_i = 1, \\
& C3: \; 0\leq i \leq N,   \\
& C4: \;  z \in \{0, 1\}. 
\label{eq:general_problem}
\end{aligned}
\end{equation}

\comment{
\begin{equation}
\begin{aligned}
\max \quad & \sum_{i=1}^{N}{Th_i}\\
\textrm{s.t.} \quad & C1: \; P_{coverage} \ge \hat{P}_{coverage} \\
& C2: \; P_{service} \ge \hat{P}_{service} \\
& C3: \; Th_i \ge D_\epsilon  \implies \beta_i = 1, \\
& C4:\; 0\leq i \leq N   \\
\label{eq:general_problem}
\end{aligned}
\end{equation}

\begin{equation}
\begin{aligned}
\max \quad & P_{coverage} \times P_{service} \\
\textrm{s.t.} \quad & C5: \; P_{coverage} < \hat{P}_{coverage} \\
& C6: \; P_{service} < \hat{P}_{service} \\
\label{eq:general_problem_2}
\end{aligned}
\end{equation}
}
We can rewrite in details the optimization problem~(\ref{eq:general_problem}) in~(\ref{eq:max_objective_function}) based on all the previous propagation formulas:

\comment{
\begin{equation}
\small
\begin{aligned}
\max_{Pr_i,\alpha_i, \beta_i, \theta_{etilt_j}} \quad & \sum_{i=1, j=c}^{N}{B_{PRB} \times N_b \times \eta_i 
\times \log_{2}(1+\frac{Pr_ij}{n_0})} \\
\textrm{s.t.} \quad 
& C1: \; \frac{\sum_{i}^{N}{\alpha_i}} {N} \ge \hat{P}_{coverage}, \\
& C2: \; \frac{\sum_{i}^{N}{\beta_i}} {N} \ge \hat{P}_{service}, \\
& C4: \; 0\leq i \leq N \;, \; 0\leq j < M, \\
&C7: \; Pr_i \ge Pr_\epsilon \implies \alpha_i = 1,\\
& C8: \; 0 \;{\rm dB} < P_{t_j}\leq 40\;{\rm dB},\\
& C9: \;  0^\circ\leq \theta_{etilt_j} \leq 14^\circ,\\
& C10: \; j \in \{\rm BS\}\smallsetminus \{\rm BS_O\},\\
\end{aligned}
\label{eq:max_objective_function}
\end{equation}

\begin{equation}
\small
\begin{aligned}
\max_{\alpha_i, \beta_i} \quad & \frac{1} {N^2} \sum_{i}^{N}{\alpha_i} \times \sum_{i}^{N}{\beta_i}\\
\textrm{s.t.} \quad 
& C4: \; 0\leq i \leq N  \\
& C5: \;  \frac{\sum_{i}^{N}{\alpha_i}} {N}  < \hat{P}_{coverage}, \\
& C6: \; \frac{\sum_{i}^{N}{\beta_i}} {N} < \hat{P}_{service}, \\
\end{aligned}
\label{eq:max_objective_function_2}
\end{equation}
}

\begin{equation}
\small
\begin{aligned}
\max_{Pr_i,\alpha_i, \beta_i, \theta_{etilt_j}} \quad & z \sum_{i=1, j=c}^{N} \left( B_{PRB} \cdot N_b \cdot \eta_i  \cdot \log_{2}\left(1+\frac{Pr_ij}{n_0}\right) \right)\\
&+ \frac{1-z}{N^2} \sum_{i=1}^{N} \alpha_i \cdot \sum_{i=1}^{N} \beta_i
\\
\textrm{subject to} \quad 
& C1: \; \rm{if} \; \mathit{P_{coverage}}\ge \mathit{\hat{P}_{coverage}} \;\;  \\ &  \;\;\;\;\;\;\;\;\rm{and} \;\;  \mathit{P_{service}} \ge \mathit{\hat{P}_{service}} \implies z = 1,  \\
& C3:  \; 0\leq i \leq N,  \\
& C4: \; 0\leq j \leq M, \\
& C5:\;  z \in \{0, 1\}, \\
&C6: \; Pr_i > Pr_\epsilon \implies \alpha_i = 1,\\
& C7: \; 0 \;{\rm dB} < P_{t_j}\leq 40\;{\rm dB},\\
& C8:\;  0^\circ\leq \theta_{etilt_j} \leq 14^\circ,\\
& C9: \; j \in \{\rm BS\}\smallsetminus \{\rm BS_O\},
\end{aligned}
\label{eq:max_objective_function}
\end{equation}
where each constraint ensures a specific condition that is detailed in Table~\ref{table:list_constraint}. Besides $Pr_ij =  P_{t_j} + G_tj(\theta_ij, \theta_{etilt_j}, \phi_ij)$ represents the received signal by the $i^{th}$ user from the $j^{th}$ BS described in~(\ref{eq:received_signal}). The parameters used to define the problem are listed in Table~\ref{table:list_parameters}.

\begin{table}[bp]
  \centering
  \begin{tabular}{!{\vrule width 1pt}p{1.5cm}|p{6.5cm}!{\vrule width 1pt}} 
    \specialrule{0.1em}{0em}{0em}
    \textbf{Constraints} & \textbf{Meaning} \\
    \specialrule{0.1em}{0em}{0em}
    C1 & If availability of coverage is higher or equal than the the coverage threshold  $\hat{P}_{coverage}$ and availability of service throughput is higher or equal than the service threshold  $\hat{P}_{service}$ then the binary variable $z = 1$.\\
    C2 & Received throughput of $i^{th}$ user should be higher than its requested throughput $D_\epsilon$ and if so the binary variable $\beta = 1$. \\
    C3 &  Range of $i^{th}$ users indexes. \\
    C4 & Range of $j^{th}$ BS indexes. \\
    C5 & Possible values of the binary variable $z$. \\
    C6 & Received signal power of $i^{th}$ user should be higher than threshold $Pr_\epsilon$ and if so the binary variable $\alpha = 1$.\\
    C7 & Range of transmit signal power. \\
    C8 & Range of electrical antenna downtilt. \\
    C9 & Range of possible values for the index $j$ includes those that belong to the set of functioning $j^{th}$ BSs that are tuned, except the indexes of BSs that are experiencing an outage state \{$BS_O$\}.\\
    \specialrule{0.1em}{0em}{0em}
  \end{tabular}
  \caption{List of constraints.}
  \label{table:list_constraint}
\end{table}

This problem falls into the category of NP-hard problems as described in~\cite{jointoptjournal}. Given the big space search and the number of radio parameters that should be tuned to find the optimal solution, traditional algorithms may struggle to provide satisfactory results as they will take a reasonable amount of time. However, RL is one promising solution to address the challenges posed by such NP-hard problems. By leveraging the power of neural networks and RL frameworks, we can learn complex decision-making policies from data and experience, enabling the network to adapt and improve users KPIs over time and thus the network resilience.

\subsection{Markov Decision Process}
\label{sec:MDP}

This problem of resilience optimization can be formulated as an MDP for the RL based solution. Formally, it is described by the tuple $(S, A, T, R, \gamma)$, where $S$ is a finite state space, $A$ is a finite action space, $T$ is the transition model, $R : S \times A \rightarrow R$ is a global reward function, and $\gamma \in [0, 1]$ is a discount factor. A deep Q-network algorithm can be used to solve this type of discrete optimization problem where:

\comment{
\begin{table}[bp]
  \centering
  \begin{tabular}{!{\vrule width 1pt}p{1.5cm}|p{6.5cm}!{\vrule width 1pt}} 
    \specialrule{0.1em}{0em}{0em}
    \textbf{Constraints} & \textbf{Meaning} \\
    \specialrule{0.1em}{0em}{0em}
    C1 & Availability of coverage should be higher than 5\%\\
    C2 & Availability of service throughput should be higher than 30\% \\
    C3 &  $w_1$ and $w_2$ positive weights for each objective in the objective function \\
    C4 & Received throughput of $i^{th}$ user higher than its requested throughput $D_\epsilon$ \\
    C5 & Range of $i^{th}$ users indexes and $j^{th}$ BS indexes \\
    C6 & Received signal power of $i^{th}$ user higher than threshold $Pr_\epsilon$ \\
    C7 & Received signal power of $i^{th}$ user from best $j^{th}$ BS antenna formula developed in equation~(\ref{eq:received_signal})\\
    C8 & Range of transmit signal power \\
    C9 & Range of electrical antenna downtilt \\
    C10 & Set of functioning $j^{th}$ BS that are tuned \\
    \specialrule{0.1em}{0em}{0em}
  \end{tabular}
  \caption{List of constraints}
  \label{table:list_parameters}
\end{table}}

\begin{itemize}
    \item \textbf{State:} represents the current network configuration of all $M$ antennas, which includes the tilt angle of each antenna and the transmit power level for each antenna. The state can be modeled by a matrix of $(M \times 2)$ size where each line corresponds to one cell antenna configuration for both transmit power and antenna tilt.
    
    \item \textbf{Action:} The action space consists of $9$ actions per cell ($M$ cells) for adjusting the tilt angle and the transmit power of each antenna simultaneously. There are 3 actions for changing the tilt angle with \{$-1^\circ$, $0^\circ$, or $+1^\circ$\} degrees, whereas there are 3 actions for changing the transmit power with \{-5dB, 0dB,~or~+5dB\}, resulting in a total of $9=3 \times 3$ possible actions per cell.
    For a three-sectored BS the action space is increased to $9^3$, and for the entire serving area of $M$ cells, the action space is $9^M$. Besides, the action status is represented by a  matrix of $(M \times 2)$ size where each line correspond to one cell antenna action for both transmit power and antenna tilt.
    We highlight the large size of the action space over the entire serving area amounting to $9^M$. Subsequently, we describe further an approach that involves solving the MDP on each BS while addressing this problem through a distributed setting, thereby, reducing the action space to $9^3$ at each BS. This approach necessitates the deployment of multiple cooperative learners agents in the entire networking area.

    \item \textbf{Next state:} is obtained by adding both state $s$ matrix and action matrix to obtain the new antennas configuration matrix $s'$.
    
    \item \textbf{Immediate reward:} we choose the immediate reward as a function that reflects the performance we want to optimize.
    For that, $R(s)$ is the reward of being in or entering a state $s$. It corresponds to the first part of the optimization function in~(\ref{eq:general_problem}) when coverage and service thresholds are satisfied, and to the second part of the optimization function when the coverage and service constraints fall below the thresholds. In other words, the state reward in~(\ref{eq:r_s}) corresponds to the total $\Th$ when $\hat{P}_{coverage}$ and $\hat{P}_{service}$ are satisfied, otherwise, the state reward corresponds to the product of $P_{coverage}$ and $P_{service}$ if the constraints fall below the thresholds.
    Besides, the immediate reward $R(s,a,s')$ for taking an action $a$ at state $s$ and resulting in state $s'$ can be expressed as the difference between the reward terms between the state $s$ and the state $s'$ (see~(\ref{eq:r_s_a})).

    \comment{
    \begin{equation}
    R(s) = \quad & \frac{w_1}{N} * \sum_{i=1,j\in BS_s}^{N}{Th_ij} + w_2 *P_{coverage}*P_{service},
    \label{eq:r_s}
    \end{equation}}

    \begin{equation}
     R(s) =  
    \begin{cases}
      \sum_{i=1,j\in BS_s}^{N}{Th_ij} & 
      \begin{array}{@{}l@{}}
        \text{if } \mathit{P_{coverage}} \ge \mathit{\hat{P}_{coverage}} \\ 
        \text{and } \mathit{P_{service}} \ge \mathit{\hat{P}_{service}}
      \end{array}
      \\
      P_{coverage} \times P_{service} & \rm{otherwise},
    \end{cases}
    \label{eq:r_s}
\end{equation}

    \begin{equation}
    \small
    \begin{split}
            R(s,a,s') =  R(s') -  R(s).
    \end{split}
        \label{eq:r_s_a}
    \end{equation}

\end{itemize}

We note that, the values of the first term are too large compared to the second term of the reward function. In fact, the term $P_{coverage} \times P_{service}$ is playing the role of a penalty in the learning process, forcing the optimization to satisfy the coverage and service constraint. Having a dynamic penalty term, instead of a static value like a zero penalty, will provide a gradient of the feedback rather than a binary or static response. This allows the learning algorithm to understand the severity of constraint violations in a more nuanced way, leading to encourage improvements or outage mitigation rather than just meeting a minimum threshold. 
For solving this MDP, the multi-agent deep Q-network DQN) algorithm is considered in the following section.

\subsection{Multi-Agent DRL}

Q-learning is a model-free RL algorithm that enables agents to learn optimal actions in Markovian domains through experiences, based on the states $s$, actions $a$ and rewards $R$, collected by the agent throughout its interactions with the environment. The mapping from environment states to actions is called policy $\pi$, which dictates the agent's behavior given a particular situation. In order to find the optimal policy $\pi^*$, Q-learning applies Policy iteration. It takes some initial policy $\pi$, and constantly evaluates an action-state value function for that policy. This value function is a measure that indicates how good each interaction with the environment is. It is defined as the expected cumulative reward for taking an action $a$ in a state $s$ and then following a policy $\pi$ (see~(\ref{eq:q_funct})). Unlike immediate rewards, the value function provides insight into the long-term desirability of interactions.
    
\begin{equation}
Q^\pi(s,a) = \mathbb{E}[\sum_{t\geq 0}^{ }{\gamma^t R_t} \vert s_0 = s,a_0=a,\pi].
\label{eq:q_funct}
\end{equation}

The Q-learning algorithm updates its policy 
following the value iteration function that is proven to converge to $Q^*$ as $i \rightarrow \infty$ in~(\ref{eq:value_iter}). It is 
based on the Bellman expectation~(\ref{eq:value_iter}) that decomposes the Q-value function into, the expected cumulative reward of the immediate reward $R$ and the Q-value function of the next state $s'$ (i.e., $s_{t+1}$ with time reference). It improves the policy by greedily taking the action that maximizes the Q-value in the future steps.

\begin{equation}
    Q_{i+1}(s,a) = \mathbb{E}_{s'\sim \epsilon} [r+ \gamma \max_{a'} Q_i^*(s',a')\vert s,a].
    \label{eq:value_iter}
\end{equation}

In real world RL problems, it is common to use non-linear function approximators to estimate the action-value function by continuously adjusting it towards the action-value function corresponding to the optimal policy, denoted as~$Q(s,a;\theta)~\approx~Q^*(s,a)$. For instance, the function approximator can be a neural network such as the DQN algorithm.

\setcounter{algorithm}{-1}
\begin{algorithm}[htp]
\small
\caption{Multi-DQN Algorithm}
\begin{algorithmic}
\For{Agent = 1, N\_agents}
\State Initialize Replay memory $D$ to capacity N
\State Initialize action-value function Q with random weights $\theta$
\State Initialize target action-value function $hat{Q}$ with weights $\theta^- = \theta$
\EndFor
\For{ episode = 1, N\_episode }
    \State Set random number of BS outage N\_BS\_off
    \State Set random index of BS outages BS\_off\_idx
    
    \For{ Agent = 1, N\_agents }
        \If{Agent not in BS\_off\_idx}
            \State Initialize sequence $s_t$ and preprocessed sequence $\phi_1$
         \For{ episode = 1, T }
         
        \State With probability $\epsilon$ select a random action $a_t$
        \State otherwise select $a_t = argmax_aQ(\phi(s_t),a:\theta)$
         \State Execute action $a_t$
         
         \For{user i = 1, N}
         \For{cell j = 1, M}
         \If{cell is working}
         \State Compute $Pr_ij$
         \EndIf
         \EndFor
         \EndFor
         \State Sort all $Pr_ij$ for each user
         \State Apply load balancing and apply cell selection 
        \State Compute immediate reward $r_t$ and observe image $x_{t+1}$
        \State Set $s_{t+1} = s_t, a_t, x_{t+1}$
        \State preprocess sequence $\phi_{t+1} = \phi(s_{t+1})$
        \State Store transition $(\phi_t, a_t, r_t, \phi_{t+1})$
        \State Sample random minibatch of transitions $(\phi_j, a_j, r_j, \phi_{j+1})$ from $D$
        
        \State Set $y_j = \begin{cases}
        r_j &  \text{if}\; s_{t + 1} \; \text{terminal}, \\
        \begin{aligned} r_j \; + \;\;\;\;\;\;\;\;\;\;\;\;\; \;\;\;\;\;\;\;\;\;\;\;\;\\ 
        \gamma \max_{a'} \hat{Q}(\phi_{j+1}, a'; \theta^-)
        \end{aligned}
         & \;\;\;\text{otherwise},
        \end{cases}$
        \State Perform a gradient descent step on $(y_j - Q(\phi_j, a_j;\theta))^2$ with
        \State respect to the network parameters $\theta$
        \State Every C steps reset $\hat{Q} = Q$ for each agent
        \EndFor
        \EndIf
    \EndFor
\EndFor
\end{algorithmic}
\label{alg:multi_dqn}
\caption{Multi-agent DQN Algorithm}
\end{algorithm}

We refer to a neural network function approximator with parameters (weights and biases) $\theta$ as a DQN. It maintains two separate Q-networks $Q(s,a;\theta)$ and $Q(s,a; \theta^-)$ with current parameters $\theta$ and old parameters $\theta^-$ respectively. The DQN agent starts by exploring randomly all the possible actions and states with some unknown reward distributions. It uses experience replay by 
storing the last $n$ experience tuples $e_t = (s_t, a_t, R_t, s_{t+1})$ into a replay memory $D = \{e_1, .... e_n\}$, and samples uniformly at random from $D$ when performing updates. During the interaction, the agent selects and executes an action according to an $\epsilon$-greedy policy using the current network updated many times per time-step, and copies the old parameters $\theta^-$ after $n$ iterations. As far as the interaction goes on, the agent needs to exploit the best action found so-far from exploration by taking sequential decisions that maximizes the cumulative reward. Having the right balance between exploration and exploitation guarantees the convergence towards the Bellman target.

\begin{table}[hbp]
\scriptsize
  \centering
  \begin{tabular}{p{3.5cm}|p{3.5cm}} 
    \toprule
    \textbf{Parameter} & \textbf{Value} \\
    \midrule
    Network type & Hexagonal \\
    Cellular layout & 21 small cell sites \\
    Inter-site distance & 300m \\
    Sectors & 3 sectors per cell \\
    Number of BSs & 7 BSs\\
    Number of outage BS & $L \in\{1,2,3,4,5\}$\\
    User distribution & Uniform random distribution\\
    Path loss & 'TR38901\_UMiv2' \\
    Number of users & 2500 users \\
    Users speed & 3-10 KM/h \\
    UT height $h_{UE}$ & 1.5 meters\\
    BS height $h_{UE}$ & 10 meters\\
    Frequency Band & 28GHz   \\
    Mechanical downtilt  & $\theta_{mtilt}$ = 0° \\
    Electrical downtilt  & $\theta_{etilt} \in$ [0°,14°] \\
    UEs Requested data rate (\textbf{$D_\epsilon$}) & 3 Mbps \\
    Number of PRB & 100\\
    $\phi_{BS}$ antenna boresight & \{0°, 120°, -120°\}\\
    Antenna max gain $G_{max}$ & 8dBi\\
    Elevation Half power beamwidth  & $\theta_{3dB}$ = 65° \\
    Azimuth Half power beamwidth  & $\phi_{3dB}$ =65° \\
    Noise per PRB & -99dBm \\
    $SSL_v$ & 30 dB \\
    $SSL_h$ & 30 dB \\
    User antenna gain $G_r$ & 0dBi \\
    User antenna type & Omnidirectionnal \\
    PRB Bandwidth & 10 MHz\\ 
    Bits per PRB & 1.4 bits per symbol\\
    $\hat{P}_{coverage}$ & 0.95\\
    $\hat{P}_{service}$ & 0.5\\
    \bottomrule
  \end{tabular}
  \caption{Network configuration on simulator.}
  \label{table:simulation_parameters}
  \vspace{-0.2cm}
\end{table}

In order to address our resilience optimization problem formulated in section~\ref{sec:probb_form} and solve its correspondent MDP from section~\ref{sec:MDP}, a centralized version using only one DQN agent is unfeasible due to the the complexity brought by the large action space. The multi-agent framework involves training multiple agents simultaneously, where each agent is executed at one BS within the network that serves $3$ cells. 
This approach is particularly well-suited to our problem domain, as the action space associated with optimizing the operation of each individual BS is vast and complex. 
Unlike a centralized approach with a single agent, that may struggle to handle the $M$ cells with an action space of $9^M$, the multi-agent DQN enables each BS to autonomously learn and adapt its behavior based on local observations over $9^3$ action space.
In each training episode, the environment randomly selects BS outages, assigning inactivity to their correspondent agents,
while enabling operational BSs and their associated active agents to adapt their strategies based on rewards. The reward for each agent is calculated using observations and propagation effects across the entire area, rather than solely on the local BS. It accounts for load balancing and maximal signal strength for an efficient cell selection for each user. Thus, this reward structure facilitates indirect communication between agents, fostering cooperative learning. This decentralized approach facilitates efficient exploration and exploitation of network dynamics, enhancing performance and resilience in the face of varying network conditions. The full pseudo algorithm is described in Algorithm~1. In the next section, we apply our resilience optimization technique on a deployment of 5G small cells network serving area to assess experimentally its capability and enhancement efficiency.

\section{Experimental Results}

In this section, we start by presenting the system configuration parameters and setups. We will first show and discuss the training performance of the proposed multi-agent DRL. Then, we present the evaluation of the resilience solution on the simulated environment on both coverage and quality of service optimization achieved by the proposed multi-agent DRL method. 

\begin{table}[tp]
\scriptsize
  \centering
  \begin{tabular}{|p{2cm}|p{6cm}|} 
    \hline
    \textbf{Term} & \textbf{Meaning} \\
    \hline
    \textbf{Covered} users & Users that receive the minimal signal strength threshold $Pr_i > -127dBm$ (see table \ref{table:RF_conditions}) \\
    \textbf{Satisfied} users & Users that receive the minimal requested data rate where their $Th_i \geq$ $D_\epsilon$ \\
    \textbf{Good} $\RSRP$ users & Users that receive at least a signal strength of -90dBm (see table \ref{table:RF_conditions})\\
    \textbf{Fair} $\RSRP$ users & Users that receive a signal strength of between -126dBm and -90dBm (see table \ref{table:RF_conditions})\\
    \textbf{Poor} $\RSRP$ users & Users that receive a signal strength of below -127dBm (see table \ref{table:RF_conditions})\\
    \textbf{BS off} & A Base Station that is experiencing an outage and is off from transmitting any signal \\
    \textbf{Coverage Availability} & Percentage of covered users in the network area \\
    \textbf{Service Availability} & Percentage of Satisfied users in the network area\\
    \textbf{Resilient Network} & Coverage availability greater than $\hat{P}_{coverage} = 95\%$ and Service availability greater than $\hat{P}_{service} = 50\%$ \\
    \hline
  \end{tabular}
  \caption{List of terminology for the experimental section.}
  \label{tab:list_terms}
  \vspace{-0.5cm}
\end{table}

\begin{figure}[bp]
\centering
 \vspace{-0.6cm}
    \begin{subfigure}[b]{0.45\textwidth}
    \centering
    \includegraphics[width=8cm, height=3.5cm]{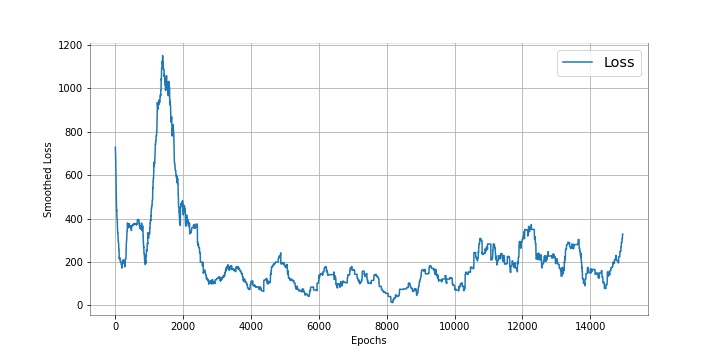}
    \caption{}
    \label{subfig:loss_DQN}
    \end{subfigure}
    \hfill
    \begin{subfigure}[b]{0.45\textwidth}
    \centering
    \includegraphics[width=8cm, height=3.5cm]{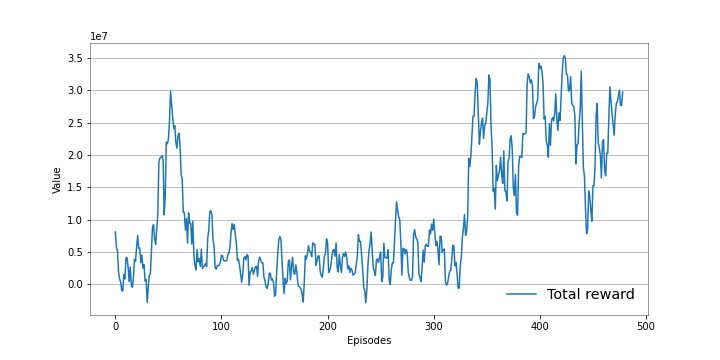}
    \caption{}
    \label{subfig:reward_DQN}
    \end{subfigure}
 \caption{(a) Single agent evolution of its smoothed loss over time - window size = 500 steps; (b) Single agent evolution of its total reward over time - window size = 5 episodes.}
 \label{fig:loss_reward_DQN}
 \vspace{-0.1cm}
\end{figure}

\begin{figure*}[!htp]
\centering
    \begin{subfigure}[b]{0.45\textwidth}
    \centering
    \includegraphics[width=7cm, height=4.2cm]{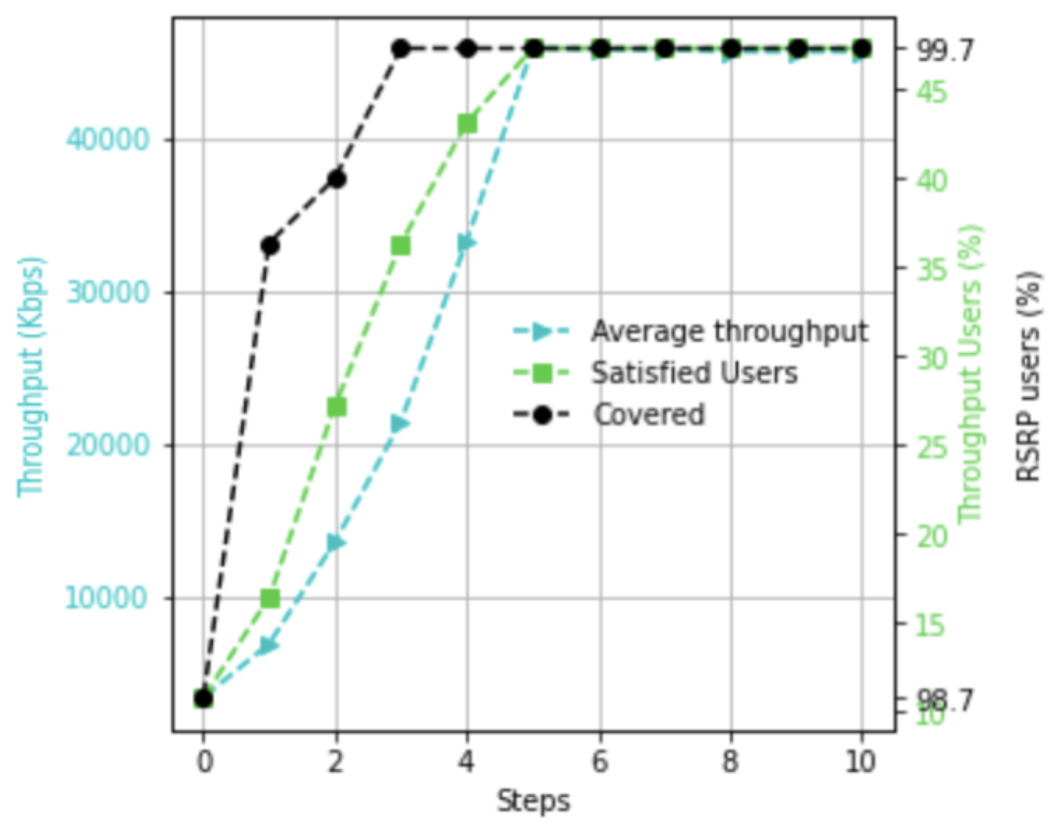}
    \caption{}
    \label{subfig:inf_2BS}
    \end{subfigure}
    \begin{subfigure}[b]{0.45\textwidth}
    \centering
    \includegraphics[width=7cm, height=4.1cm]{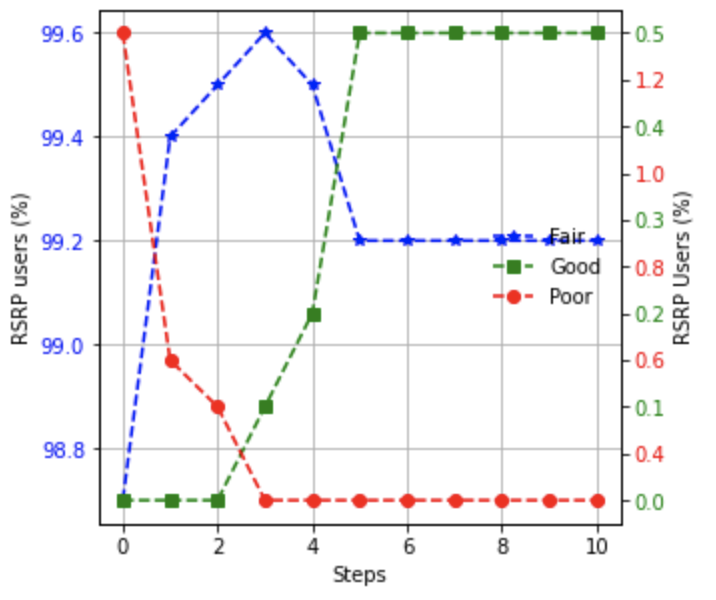}
    \caption{}
    \label{subfig:RSRP_2BS}
    \end{subfigure}
    \begin{subfigure}[b]{0.45\textwidth}
    \centering
    \includegraphics[width=7cm, height=4.2cm]{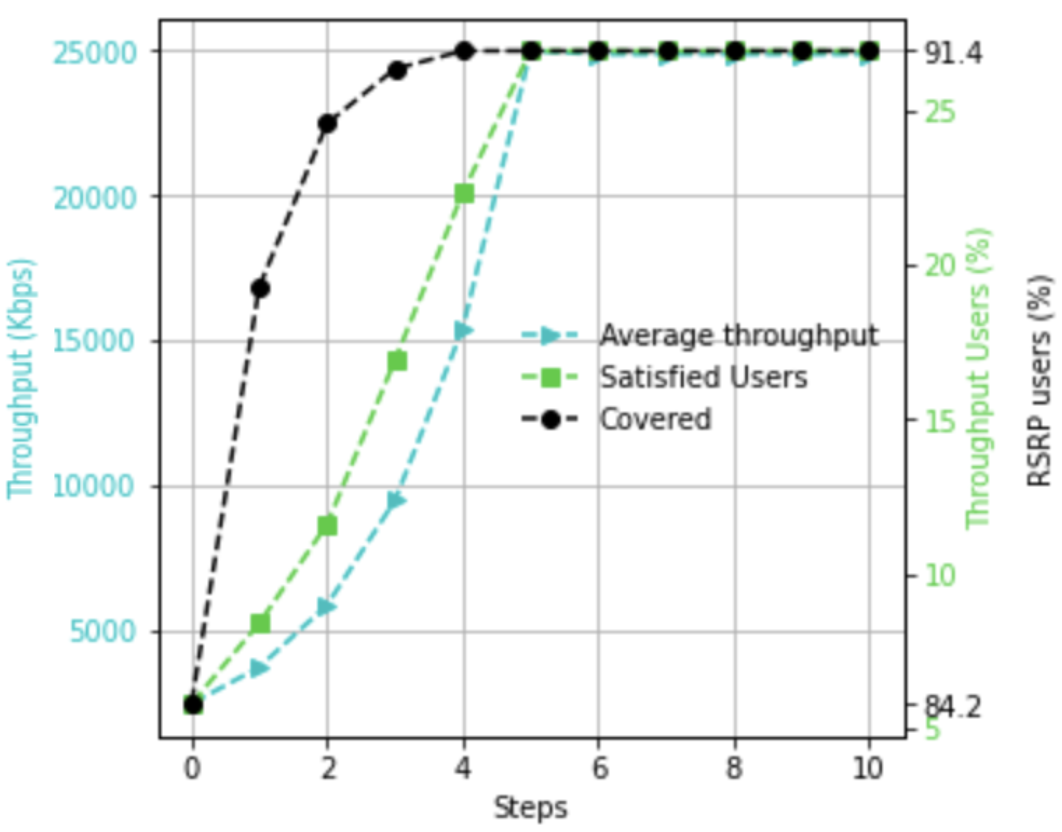}
    \caption{}
    \label{subfig:inf_3BS}
    \end{subfigure}
    \begin{subfigure}[b]{0.45\textwidth}
    \centering
    \includegraphics[width=6.5cm, height=4.1cm]{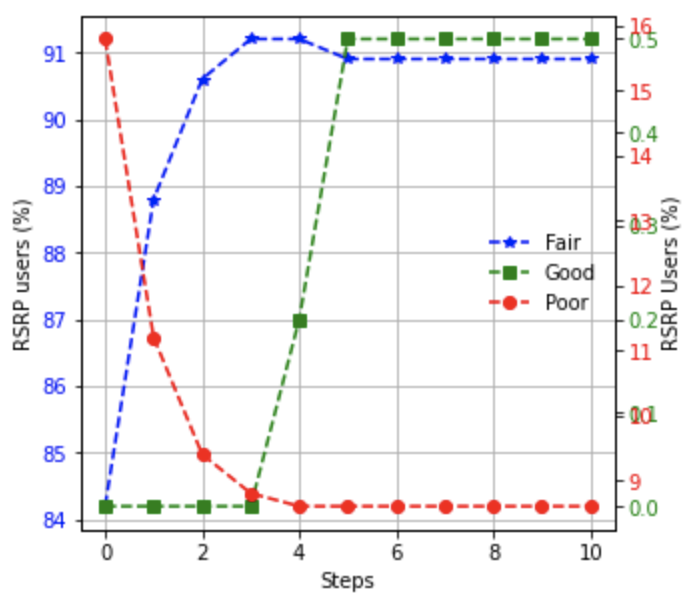}
    \caption{}
    \label{subfig:RSRP_3BS}
    \end{subfigure}
    \begin{subfigure}[b]{0.45\textwidth}
    \centering
    \includegraphics[width=7cm, height=4.1cm]{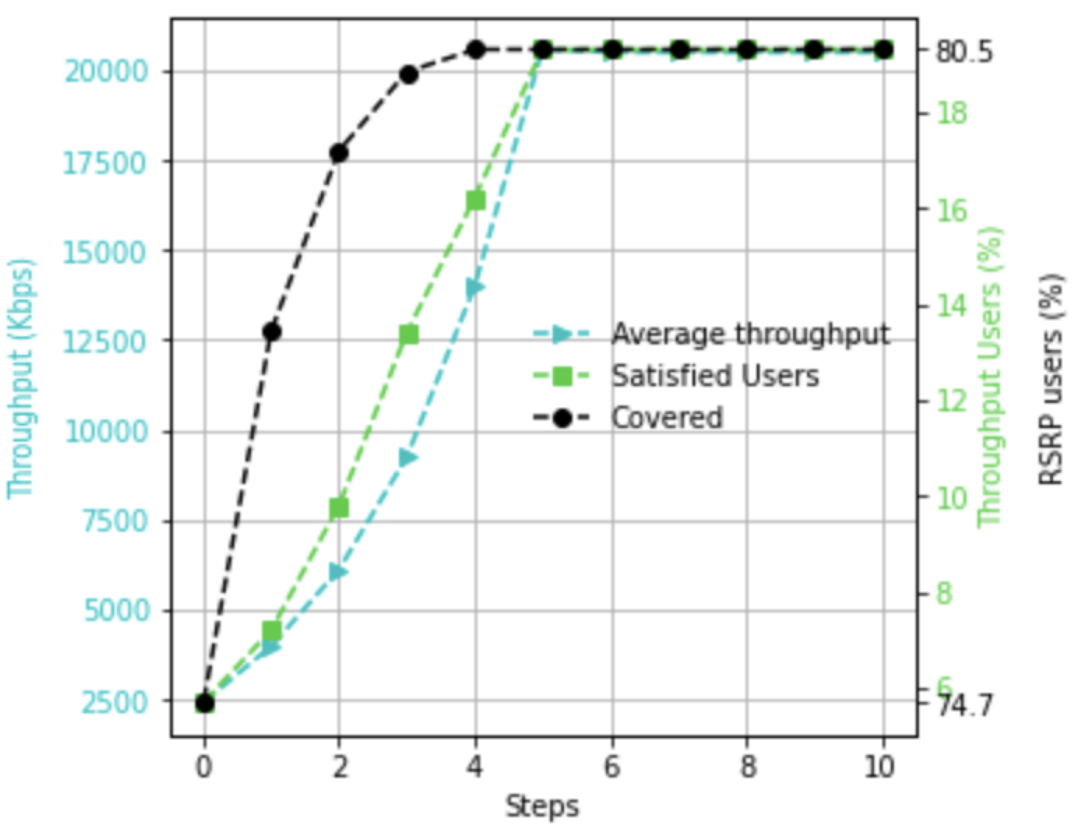}
    \caption{}
    \label{subfig:inf_4BS}
    \end{subfigure}
    \begin{subfigure}[b]{0.45\textwidth}
    \centering
    \includegraphics[width=6.5cm, height=4.1cm]{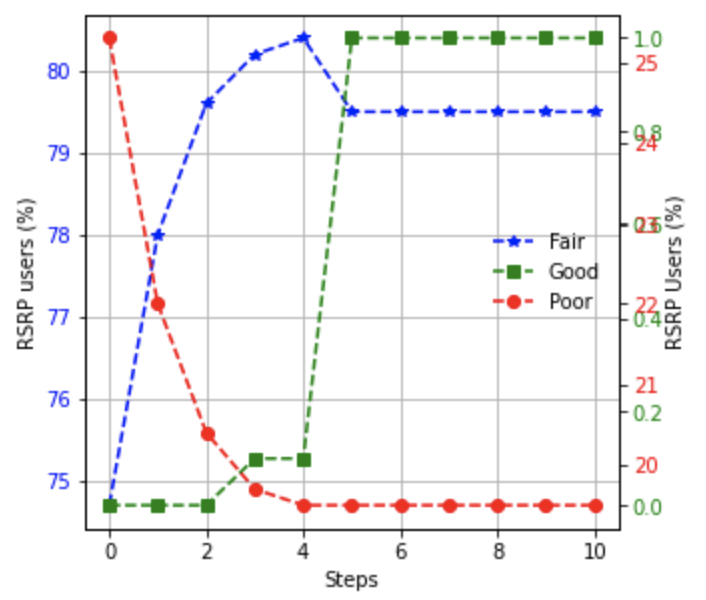}
    \caption{}
    \label{subfig:RSRP_4BS}
    \end{subfigure}
 \caption{Inference of multi-agent DQN when (a,b) 2 BS outage, (c,d) 3 BS outage and (e,f) 4 BS outage. 
 The estimation of: 
 (a,c,e) Percentage of satisfied and covered users service with the average $\Th$ evolution, and (b,d,f) The percentage of $\RSRP$ Good, Fair, Poor users evolution.}
 \label{fig:DQN_inference_3cases}
  \vspace{-0.5cm}
\end{figure*}

\subsection{Simulation setup}

To simulate the 5G RAN based on 3GPP specifications, we employ a full dynamic system tool. We set up a baseline reference scenario that consists of 21 gNBs (i.e., BSs) having an inter-site distance of 300m. This layout derives the presence of 7 BSs, each deploying a single DQN agent, resulting in a total of seven distributed RL agents. To evaluate the resilience solution on the simulator, we consider 2,500 testing points uniformly distributed across the service area, each representing a connected user in the network.
The detailed network configuration parameters are listed in Table~\ref{table:simulation_parameters}.

To simulate a hardware failure in the network, at some point in the simulation, the antenna gain of the selected BSs is attenuated to -100dBi that leads to a cell outage in a network.
We extract path loss measurements from the simulator to obtain accurate attenuation estimates at each testing user point and utilize them in the propagation formula.
For the training, 
we deliberately vary the degrees of stress and failure scenarios, by randomly selecting up to five BSs for outage simulation. The number and the index of BSs are randomly selected during the training. We vary the number of BS outage $L$ from 1 to 5.  The case of all BSs outage except one (i.e., 6 BSs off) is not considered since we assume that one BS alone is incapable of recovering all the users of the network area. Besides, the case of all BS outage (i.e., 7 BS off) is also not considered because no BS is left for the outage mitigation. By considering these different situations during training, this enables the DRL models to learn adverse events on the underlying network environment and evaluate the overall performance and resilience throughout this study.


\begin{figure*}[htp]
\centering
    \begin{subfigure}[b]{0.45\textwidth}
    \centering
    \includegraphics[width=7cm, height=4.3cm]{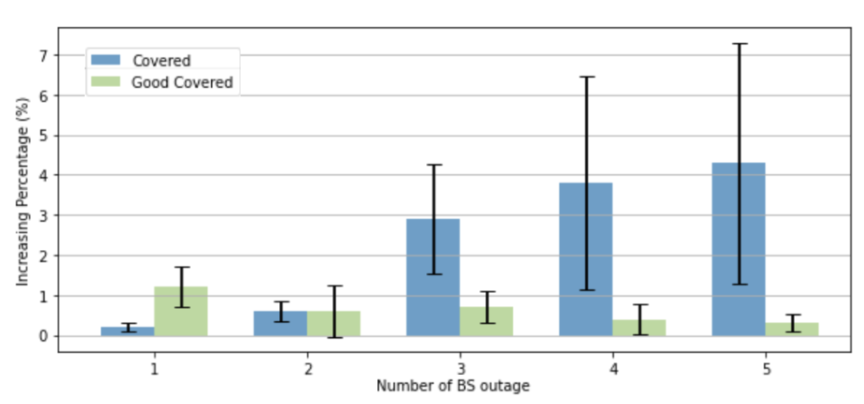}
    \caption{}
    \label{subfig:symmary_res_coverage}
    \end{subfigure}
    \hspace{0.5em}
    \begin{subfigure}[b]{0.45\textwidth}
    \centering
    \includegraphics[width=7cm, height=4.1cm]{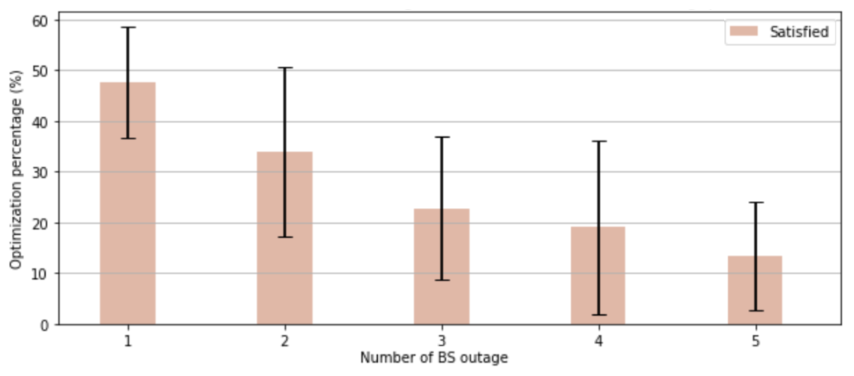}
    \caption{}
    \label{subfig:symmary_res_service}
    \end{subfigure}
 \caption{Estimation of average performance increase over multiple inferences for: (a) Coverage availability through  the percentage of covered and good $\RSRP$ users increase (b) Service availability through the percentage of satisfied users receiving the requested Th.}
 \label{fig:symmary_results}
 \vspace{-0.5cm}
\end{figure*}

The evaluation focuses on tracking the number of covered users and the satisfaction level of users with their service. Additionally, we monitor the evolution of average $\RSRP$ and $\Th$ metrics under multiple network conditions.
First, we discuss the loss and total reward evolution for one DQN agent as an example (see Fig. \ref{fig:loss_reward_DQN}), and perform multiple inferences with the proposed multi-agent DQN varying the number of BS outages from 1 to 5. 
We present the performance estimation evolution on all cases in Fig.~\ref{fig:symmary_results}, and illustrate the network behaviour in 3 cases of BS outages on Fig.~\ref{fig:DQN_inference_3cases}. Then, we evaluate the proposed multi-agent DQN solution in a dynamic scenario simulated using the SONTool simulator~\cite{SONTool1}, developed and maintained up-to-date on Matlab. The downlink principles of the simulator are presented in~\cite{SONTool1} and uplink modeling aspects in~\cite{SONTool2}. Subsequently, we deploy our proposed resilience solution based on multi-agent DQN in the simulator to trigger it dynamically when an outage state is detected, and track the network performance on a realistic environment propagation conditions. 
This standardized simulator will closely mirror reality and capture essential aspects of network behavior while reflecting the performance of our resilience strategy. We evaluate the resilience solution in scenarios where the number of BSs varies between two distinct sets: $L \in \{1,2,3\}$ and $L \in \{4,5\}$. To assess the benefit of our resilience solution based on the multi-agent DQN, we compare it with a benchmark solution that consist in changing only neighbouring BSs (i.e., local optimization and centralized compensation like in Fig.\ref{fig:approach}), as well as with no resilience solution settings in the case of 1 BS off.
The list of terminology that will be used throughout this experimental section is listed in table~\ref{tab:list_terms}. 
We note that other experiments have been conducted using multi-agent proximal policy optimization algorithm, which yielded similar results, emphasizing the effectiveness of the distributed solution. However, we show only the results with multi-agent DQN for the seek of simplicity.

This comprehensive analysis and discussion will allow us to gauge the effectiveness of the resilience solution across different BS configurations and provide valuable insights into its impact on network performance and user experience.


\subsection{Training results}

\textit{- Single agent DQN training performance:} We train each DQN model for 1,000 episodes, at each episode, a 500 time steps are taken by the agent for learning a single path of the Q-value. In our investigation of the DQN training process, 
we closely monitored the evolution of loss and total reward over successive training iterations. 
We plot one agent loss and total reward evolution over the total number of episodes in Fig.~\ref{fig:loss_reward_DQN}. As the DQN model begins its training journey, we observe a rapid decline in loss accompanied by a notable increase in total reward. This initial phase of training demonstrated the network's ability to swiftly learn and adapt to the environment, effectively optimizing its policy to maximize cumulative reward. However, throughout the training, we noticed an instability in the evolution of total reward which reflects the known behaviour of the DQN. Despite this, the DQN continued to exhibit a steady improvement in loss roughly approaching  convergence. These observations indicate that our DQN model successfully learn to face the underlying network problems to improve the overall performance.

\textit{- Multi-agent DQN inference performance:} 
To analyse the performance and resilience enhancement of our model, we conduct multiple inferences with 10 actions with one action per step of the multi-agent DQN under scenarios involving~1~to~5 outages.
In Fig.~\ref{fig:DQN_inference_3cases},  we plot in the one hand the coverage enhancement indicators: coverage availability evolution (i.e., covered users percentage) and the percentage of good, fair, and poor $\RSRP$ users. 
In the other hand we plot the data-rate service indicators: service availability evolution (i.e., satisfied users percentage) as well as the average throughput $\Th$. As an example we focus on scenarios when number of BS outage $L \in$ \{2, 3, 4\} outages. Since the case of 1 BSs outage is highly similar to 2 BSs outage and the case of 5 BSs outage is similar to 4 BSs outage. Further in Fig.~\ref{fig:symmary_results}, we provide a summary of coverage and service availability performance enhancement (average and standard deviation) in all outage cases. 

\begin{figure*}[htp]
\centering
    \begin{subfigure}[b]{0.45\textwidth}
    \centering
    \includegraphics[width=6.5cm, height=3.8cm]{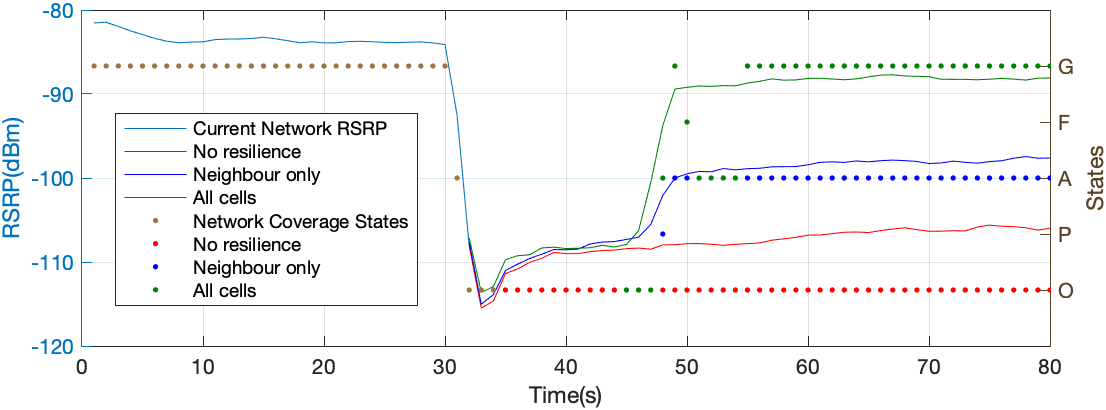}
    \caption{}
    \label{subfig:1BS_RSRP}
    \end{subfigure}
    \begin{subfigure}[b]{0.45\textwidth}
    \centering
    \includegraphics[width=7cm, height=4cm]{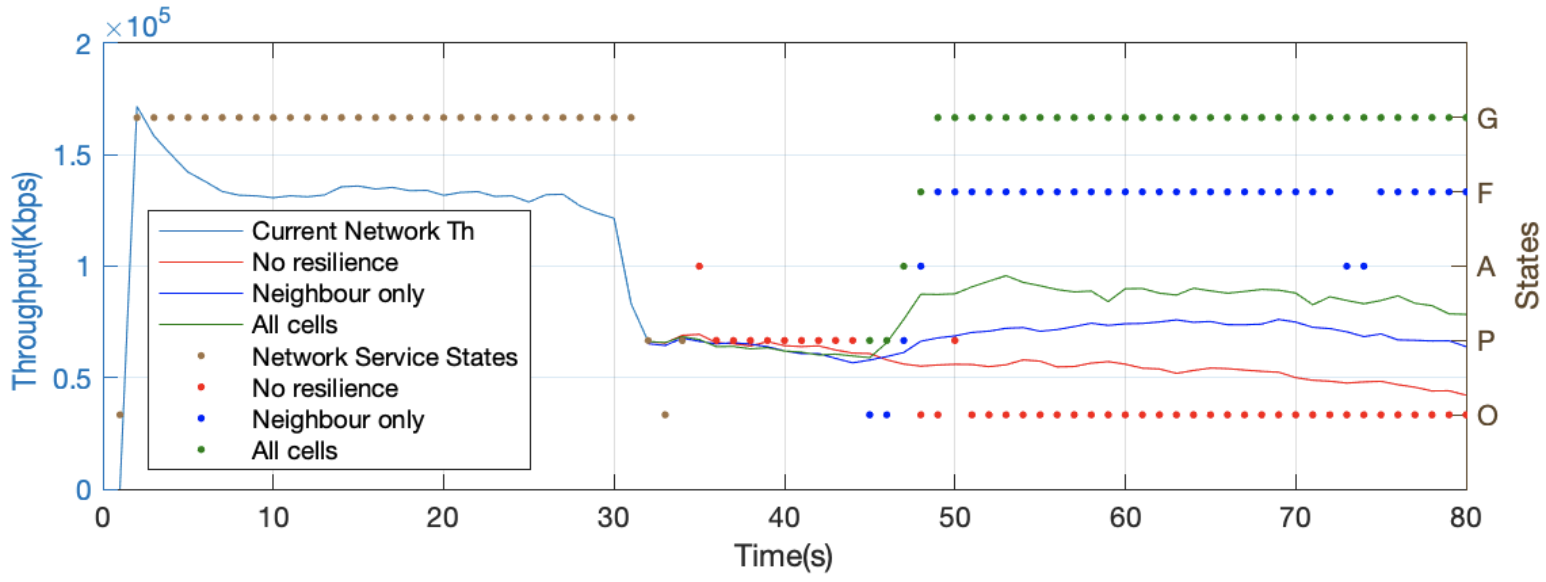}
    \caption{}
    \label{subfig:1BS_Th}
    \end{subfigure}
    \begin{subfigure}[b]{0.45\textwidth}
    \centering
    \includegraphics[width=6.5cm, height=3.8cm]{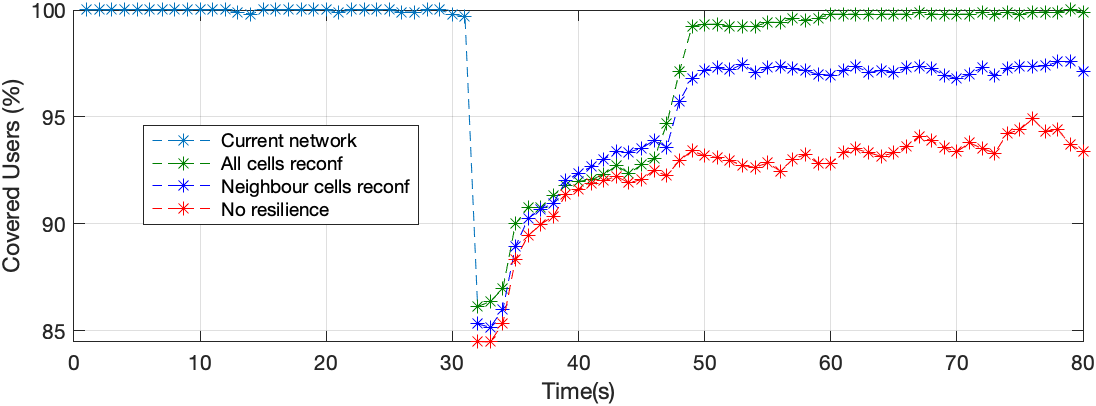}
    \caption{}
    \label{subfig:1BS_coverage}
    \end{subfigure}
    \begin{subfigure}[b]{0.45\textwidth}
    \centering
    \includegraphics[width=6.5cm, height=3.8cm]{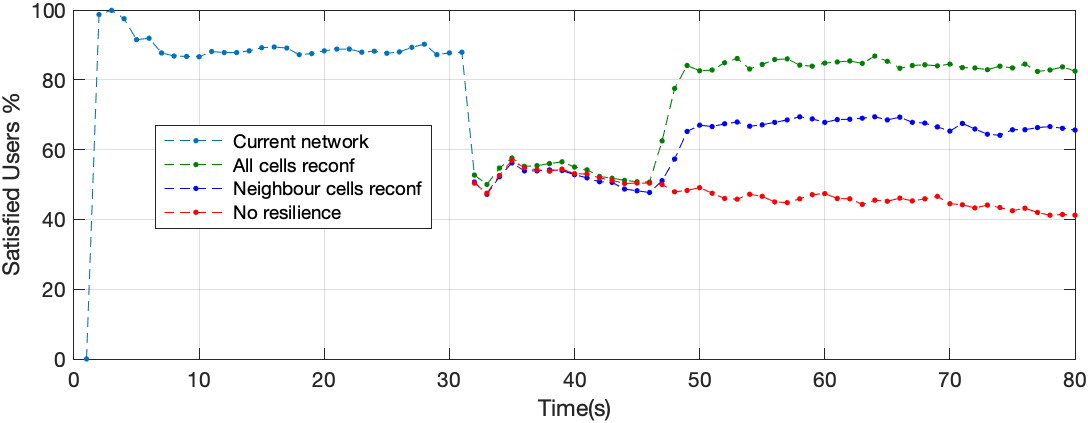}
    \caption{}
    \label{subfig:1BS_service}
    \end{subfigure}
 \caption{
 Network performance evolution of the current network in case of 1 BS outage with: no Resilience, neighbouring reconfiguration only, and multi-Agent DQN-based full cell Reconfiguration for: (a) Average received user signal strength ($\RSRP$) (b) Average $\Th$ (c) Percentage of covered users (d) Percentage of satisfied users.}
 \label{fig:1BS_summary}
  \vspace{-0.5cm}
\end{figure*}

\textit{- Multi-agent DQN inference examples:}
We notice in Fig.~\ref{fig:DQN_inference_3cases} that in all cases, our solution succeeds to improve the overall network performance, which implies improving coverage and service availability and by extension reduce network downtime and optimize the network resilience.
For example, in Fig.~\ref{subfig:inf_2BS}, Fig.~\ref{subfig:inf_3BS} and Fig.~\ref{subfig:inf_4BS} the percentage of covered users is improved by up to 99.7\% for 2 BSs outage, 91\% for 3 BSs outage and by up to 80\% in case of 4 BSs outage. The last case is not sufficient to pull out the network status from the outage state as the percentage of uncovered users is still greater than 5\%. However, the operator can involve other surrounding cells which are not showed in the studied zone to help achieving a better performance enhancement.
In addition to these observations, the percentage of satisfied users is improved by up to 
50\% for 2 BSs outage, 30\% for 3 BSs outage and by up to 20\% in case of 4 BSs outage.
These lower percentages compared to coverage can be attributed to an insufficient received user signal strength $\RSRP$, resulting in lower service quality expected during severe coverage issues. Besides, in all cases the average $\Th$ is successfully improved, as learnt by the DQN models. As for coverage through $\RSRP$ users percentage indicators in Fig.~\ref{subfig:RSRP_2BS}, Fig.~\ref{subfig:RSRP_3BS} and Fig.~\ref{subfig:RSRP_4BS}, we notice that at the early stages of action steps, the poor users percentage starts to decrease while the fair users percentage starts to increase. Later on, the fair users percentage decreases while the good users percentage increases and the poor users remains at its lowest value. This aligns with the objective of our multi-objective optimization algorithm, which aims to enhance not only availability but also to optimize the average performance and quality of service during outages.

\textit{- Multi-agent DQN inference results summary:} The summary of both coverage and service enhancement estimation statistics (median and standard deviation) are provided in Fig.~\ref{fig:symmary_results}. 
We perform over 20 inferences with 10 action steps in each BSs outage case. We compute the increase percentage difference obtained within each inference for both covered and satisfied users, as well as for good $\RSRP$ users percentage increase. We derive the median and standard deviation for each outage case. Specifically, we notice that in cases of 1, 2 and 3 BSs outage the coverage is increased with 1-4\% only while it achieves 5-7\% in cases of 4 and 5 BSs outage.  
However, the estimated increase of good $\RSRP$ users percentage does not exceed 2\% improvement in all cases. In fact, the percentage of covered users' improvement increases with the number of BS outages, while the improvement of good $\RSRP$ users decreases as the number of BSs outage increases. This phenomenon is explained by the fact that with more BSs outage, there are more uncovered users, and in cases of severe degradation (4 and 5 BSs outage), our algorithm
gives priority to recovering more users than improving the quality of service for some users.
This observation is consistent with the goal of our multi-objective optimization algorithm and underscores the adaptability of our proposed solution.
As for the satisfied users improvement, we reach up to 40-60\% of quality of service improvement in best cases (1, 2, and 3 BSs outage) and reach up to 20-30\% in worst cases (4 and 5 BSs outage). This percentage decreases as the number of BS outages increases. In fact, under severe degradations, more uncovered users can hardly be recovered with a good $\RSRP$ as the solicited compensating BSs are quite far, and subsequently, achieving requested $\Th$ becomes more challenging to ensure good quality of service, which explains this trend. In these cases, our algorithm prioritizes 
the availability of minimal conditions for the majority of users.

These estimation of performance and resilience enhancement reflect the success of our proposed solution in improving the network resilience. It also underscores the fact that the algorithm based on DRL is well suited for seeking both availability (i.e., resilience) and quality of service improvement under abnormal network conditions. Nevertheless, these discussed statics are only estimations obtained with propagation formula and model inference, we aim to evaluate our proposed solution in a real-like and dynamic network to asses the effectiveness of the solution and its impact on network resiliency. The next section is dedicated for the evaluation on the standardized dynamic simulator.

\begin{figure*}[!htp]
\centering
    \begin{subfigure}[b]{0.45\textwidth}
    \centering
    \includegraphics[width=7cm, height=3.8cm]{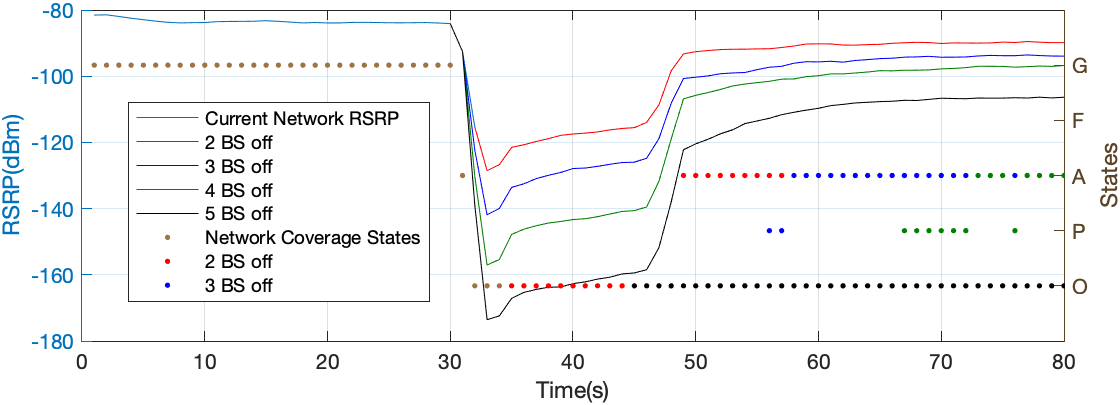}
    \caption{}
    \label{subfig:5BS_RSRP}
    \end{subfigure}
    \hspace{0.5em}
    \begin{subfigure}[b]{0.45\textwidth}
    \centering
    \includegraphics[width=7.3cm, height=3.8cm]{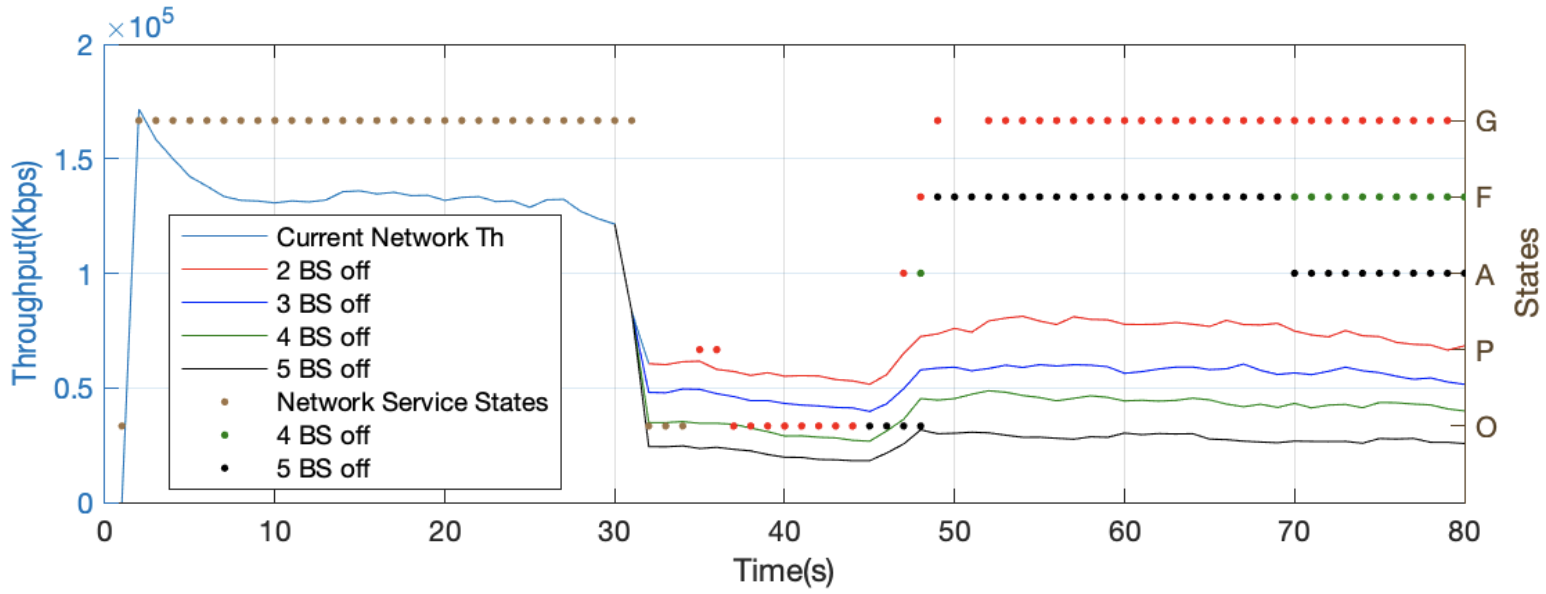}
    \caption{}
    \label{subfig:5BS_Th}
    \end{subfigure}
    \hspace{0.5em}
    \begin{subfigure}[b]{0.45\textwidth}
    \centering
    \includegraphics[width=6.8cm, height=3.8cm]{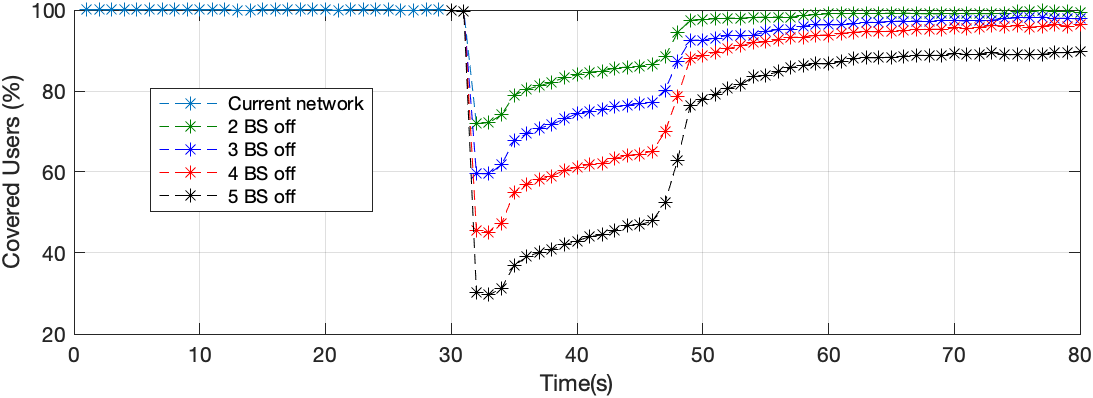}
    \caption{}
    \label{subfig:5BS_coverage}
    \end{subfigure}
    \hspace{0.5em}
    \begin{subfigure}[b]{0.45\textwidth}
    \centering
    \includegraphics[width=6.8cm, height=3.8cm]{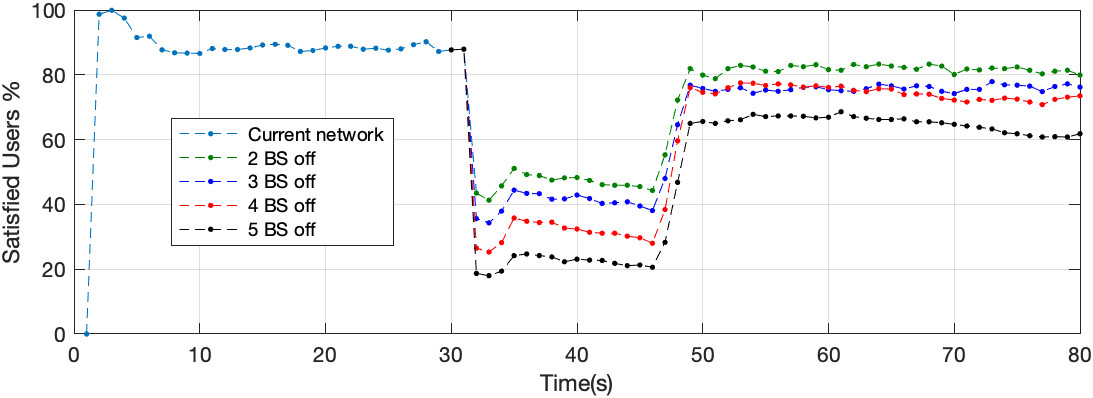}
    \caption{}
    \label{subfig:5BS_service}
    \end{subfigure}
 \caption{
 Network performance evolution of the current network in case of 2 to 5 BS outages with the Multi-Agent DQN-based full cell reconfiguration for: (a) Average Received user signal strength ($\RSRP$) (b) Average throughput (c) Percentage of covered users (d) Percentage of satisfied users.}
 \label{fig:5BS_results}
  \vspace{-0.5cm}
\end{figure*}

\subsection{Evaluation results on simulator}

At the simulator, we set a network with the configuration listed in Table~\ref{table:simulation_parameters}. We conduct different outage conditions by up to 5 BSs outages and track the current network behaviour in terms of average users $\RSRP$ and users $\Th$, as well as the percentage of covered and satisfied users. We simulate the network for 80 seconds long and stress the network at the second 30-th. We trigger dynamically the optimization solution 5 seconds after detecting an outage state. We make up to 5 actions reconfigurations on the cell sites antennas, by sending signaling information to the operational BSs to modify the electrical tilt and the transmit signal dynamically.
During the simulation, we collect $\RSRP$ and user $\Th$ at each user for a granularity of 1s. 
In Fig.~\ref{fig:1BS_summary}, we start by studying the scenario of 1 BS outage. We respectively compare the following three cases: when no resilience solution is triggered, when a conventional solution that consists of reconfiguring only neighbouring BSs as usually assumed in the literature is triggered (i.e., local optimization), and when our proposed multi-agent DRL solution is triggered to reconfigure the entire network zone. Then, we evaluate the performance of the multi-agent DRL on the cases of 2, 3, 4 and 5 BSs outage in Fig.~\ref{fig:5BS_results}.

The dynamic behaviour of the network shown in Fig.~\ref{fig:1BS_summary} reveals that our method (in green) succeeds in restoring the initial average $\RSRP$ 
(Fig.~\ref{subfig:1BS_RSRP}) and recovering back most of the users in the outage area (Fig.~\ref{subfig:1BS_coverage}) at the same level of before outage occurrence (Good coverage and service states in green points). Unlike, the conventional methods of only neighbouring cells reconfiguration (in blue) that restore back the network state to only acceptable level for coverage and fair state for service (blue points). However, the average $\Th$ is not increased at the same level as before outage, because the simulator takes into consideration different interference levels and attenuation parameters not taken into account in our propagation formulas during the DQN training. Despite this observation, the service availability is successfully restored at the requested level. In fact, some users may experience some quality of service attenuation but they still access the service with the requested levels and hence the resilience of network in terms of service is still achieved. Finally, the Fig.~\ref{fig:5BS_results} depicts the comparison of performance and resilience enhancement using our proposed multi-agent DQN approach in cases of 2, 3, 4 and 5 BSs outage. As noted before in the discussion of Fig.~\ref{fig:symmary_results}, despite the success of the proposed approach in improving network overall performance along with satisfying coverage and service availability, the mitigation technique of outages through cell reconfiguration is limited with the degree of severeness of the outage. Hence, for extreme cases or disasters, a special branch of resilience disaster study exists for deploying redundant networking equipment and BSs, especially for critical networks.

To sum up, our proposed multi-agent DQN
succeeds in learning and improving the average user received signal strength $\RSRP$ and its correspondent user $\Th$ under various network outage scenarios. It also helps to increase the percentage of covered and satisfied users, which means it enhances the coverage and service availability, resulting in improving the network resilience. This method outperforms conventional methods of only direct neighbouring cells reconfiguration and local optimization. In a broader context, the method leads to a reduction of outage downtime and increase system up-time. 
Therefore, improving the availability at time $t$ will improve the resilience of a network over a period of time. This global optimization of resilience will allow operators targeting focused network zones that experience outage such as indoor or outdoor areas (see Fig.~\ref{fig:beamforming_antenna}), leading to a more efficient resource allocation and network management. This approach could involve deploying adaptive resilience strategies across various network environments, each tailored to meet specific demands. This method have shown it's limitation in case of severe degradation and with achieving good quality of service optimization in specific outage cases, since increasing the signal transmit power of compensating BSs may not effectively mitigate the impact of severe degradation on user experience, as other factors such as interference and network congestion can play a significant role in adding complexity to the network management, especially in dense networks. 
Finally, by performing the resilience analysis developed in section~\ref{sec:analysis} and previous paper~\cite{my_paper2} in a loop along with the optimization, it will
allow to evaluate the network performance in real-time
using an end-to-end strategy.

\section{Conclusion}

In this paper, we presented a resilience optimization solution based on multi-agent deep Q-networks, in a scenario where one or multiple base stations can face outages. To optimize resilience, we defined a multi-objective reward based on maximizing total throughput while simultaneously considering enhancements in coverage and service availability. To assess the effectiveness of our proposed solution, we conducted extensive simulations, running the algorithm multiple times under varying conditions of BSs outage.  The results show that the proposed solution succeeds in improving network overall performance, and by extension the resilience of the network. This approach that considers the network zone as a cohesive entity during global optimization, can outperform traditional mitigation techniques that consist in reconfiguring only local and neighbouring cell sites of the outage area. 

Future work could involve integrating other performance indicators into the optimization, such as reducing latency and improving cell load. Additionally, an extension for a wider network area could be envisaged, with a dynamic  incorporation of the number of users. Additionally, a study could be conducted to apply the optimization to specific network areas, such as indoor or outdoor environments. Finally, an application with beamforming can also be considered and a heterogenous network resilience optimization between both macro cells and small cells can also be investigated.


\small\bibliography{references}

\newpage

\section{Biography Section}

\vspace{-33pt}
\begin{IEEEbiographynophoto}{Soumeya KAADA} is a Ph.D. candidate at University of Rennes 1 and Inria lab, Brittny, France. She worked as a research engineer during her Ph.D. at Nokia Bell Labs France.
She obtained her Master's degree in Data \& Knowledge from Telecom Paristech School and University Paris Saclay in 2019. In 2017, she received the Valedictorian title upon graduating with a Bachelor's degree in Networking and Telecommunications from USTHB University, Algeria. Her major interests are reinforcement learning, deep learning and optimization for communication and computational systems.
\end{IEEEbiographynophoto}

\vspace{-25pt}
\begin{IEEEbiographynophoto}{Dinh-Hieu Tran} (S'20) was born and grew up in Gia Lai, Vietnam. He is a senior research specialist 5G+ at Nokia, France. He finished his M.Sc degree in Electronics and Computer Engineering from Hongik University, Korea, in 2017, and the Ph.D. degree at the University of Luxembourg in 2022. His major interests include non-terrestrial-network, wireless networks. In 2016, he received the Hongik Rector Award for his excellence during his master's study at Hongik University. He was a co-recipient of the IS3C 2016 best paper award. In 2022, he was nominated for the "FNR Outstanding Thesis Award" and won the award "Excellent thesis award 2022 in Doctoral School Science and Engineering".
\end{IEEEbiographynophoto}

\vspace{-25pt}
\begin{IEEEbiographynophoto}{Nguyen Van Huynh}
(Member, IEEE) received the Ph.D. Degree in Electrical and Computer Engineering from the University of Technology Sydney (UTS), Australia in 2022. He is currently a Lecturer at the Department of Electrical Engineering and Electronics, University of Liverpool (UoL), United Kingdom. Before joining UoL, he was a Postdoctoral Research Associate in the Department of Electrical and Electronic Engineering, Imperial College London, United Kingdom. His research interests include mobile computing, 5G/6G, IoT, and machine learning.
\end{IEEEbiographynophoto}

\vspace{-25pt}
\begin{IEEEbiographynophoto}{Marie-Line Alberi Morel}
is a senior researcher at NOKIA Bell Labs. She received a Ph.D. degree in the signal processing field in 2001 from University Paris-Saclay, France. Currently, her research includes machine learning for 5G+/6G networks, deep learning, time series classification and generative models. She serves as associate professor and master's degree program head of telecom program in Gustave Eiffel University (Paris, France).  Her past works has dealt with QoE for 5G network services, super resolution, linear video coding and superposed coding for LTE and mobile broadcast network.
\end{IEEEbiographynophoto}

\vspace{-25pt}
\begin{IEEEbiographynophoto}{Sofiene Jelassi} is an assistant professor at the University of Rennes, and a researcher at IRISA Lab in Rennes, Brittny, France. He received his Ph.D. degree from Paris Sorbonne University, Paris France in 2010. His research work concerns delay-sensitive network analytics, network and system virtualization, 5G RAN (Radio Access Network) resiliency, and energy-efficient video communications. His is a contributor to more than 20 publications published in international conferences and journals. 
\end{IEEEbiographynophoto}

\vspace{-25pt}
\begin{IEEEbiographynophoto}{Gerardo Rubino}
received the Ph.D. degree in computer science and the
Habilitation degree from the University of Rennes 1 in 1985 and 1995, respectively. He is a Senior Researcher with INRIA (the French National Institute for Research in Computer Science and Control). His research interests lie in quantitative analysis
of computer and communication systems, mainly using probabilistic models. He also works on the quantitative evaluation of perceptual quality of multimedia communications over the Internet. With Bruno Tuffin, he co-edited the book Rare Event Simulation Using Monte Carlo Methods (Wiley, 2009), and co-authored several of its chapters. He has published more than 200 papers in journals and conference proceedings, in several fields of applied mathematics and computer science, and has performed various editorial tasks and managing activities in research. Pr. Rubino is a member of the IFIP WG 7.3.
\end{IEEEbiographynophoto}

\vfill

\end{document}